\documentclass{article}
\usepackage[preprint,nonatbib]{include/neurips2024/neurips_2024}

\usepackage[utf8]{inputenc} 
\usepackage[T1]{fontenc}    
\usepackage{hyperref}       
\usepackage{url}            
\usepackage{booktabs}       
\usepackage{amsfonts}       
\usepackage{nicefrac}       
\usepackage{microtype}      
\usepackage{xcolor}         
\usepackage{wrapfig}
\usepackage{graphicx}

\usepackage{caption}
\usepackage{subcaption}
\usepackage{booktabs}
\usepackage{multirow}
\usepackage{colortbl}
\usepackage{amsmath}
\usepackage{amssymb}
\usepackage{mathtools}
\usepackage{amsthm}

\usepackage{inconsolata}
\usepackage{listings}
\usepackage{algorithm}
\usepackage{algpseudocode}

\hypersetup{colorlinks=true}
\newcommand{\tb}[1]{\textbf{#1}}
\definecolor{orange2}{HTML}{ffd8a8}
\definecolor{violet2}{HTML}{d0bfff}
\definecolor{grey2}{HTML}{e9ecef}
\definecolor{indigo1}{HTML}{dbe4ff}
\definecolor{indigo2}{HTML}{bac8ff}

\title{A self-supervised framework\\for learning whole slide representations}

\author{%
Xinhai Hou\textsuperscript{*1}\quad
Cheng Jiang\textsuperscript{*1}\quad
Akhil Kondepudi\textsuperscript{1}\quad
Yiwei Lyu\textsuperscript{1}\quad\\\textbf{
Asadur Chowdury\textsuperscript{1}\quad
Honglak Lee\textsuperscript{1}\quad
Todd Hollon\textsuperscript{1}}\\[1ex]
\textsuperscript{1}University of Michigan\quad
\textsuperscript{*}Equal Contribution\\[1ex]
\texttt{\{xinhaih, chengjia, tocho\}@umich.edu}
}

\begin{document}

\maketitle

\begin{abstract}
Whole slide imaging is fundamental to biomedical microscopy and computational pathology. Previously, learning representations for gigapixel-sized whole slide images (WSIs) has relied on multiple instance learning with weak labels, which do not annotate the diverse morphologic features and spatial heterogeneity of WSIs. A high-quality self-supervised learning method for WSIs would provide transferable visual representations for downstream computational pathology tasks, without the need for dense annotations. We present \emph{Slide Pre-trained Transformers (SPT)} for gigapixel-scale self-supervision of WSIs. Treating WSI patches as tokens, SPT combines data transformation strategies from language and vision modeling into a general and unified framework to generate views of WSIs for self-supervised pretraining. SPT leverages the inherent regional heterogeneity, histologic feature variability, and information redundancy within WSIs to learn high-quality whole slide representations. We benchmark SPT visual representations on five diagnostic tasks across three biomedical microscopy datasets. SPT significantly outperforms baselines for histopathologic diagnosis, cancer subtyping, and genetic mutation prediction. Finally, we demonstrate that SPT consistently improves whole slide representations when using off-the-shelf, in-domain, and foundational patch encoders for whole slide multiple instance learning.
\end{abstract}

\begin{figure}[h!]
    \centering
    \includegraphics[width=\columnwidth]{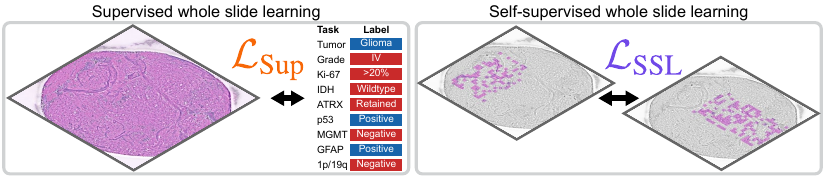}
    \caption{\textbf{Self-supervised whole slide learning.} Previous work in computational pathology relies on multiple instance learning with weak supervision from slide or patient-level labels to learn whole slide representations \cite{zhu2017wsisa, ilse2018attention, lu2021data, shao2021transmil, chen2022scaling, javed2022additive, chen2024towards}. We present a self-supervised framework for learning whole slide representations, called \emph{Slide Pre-trained Transformers (SPT)}, by combining data transformations from vision and language modeling to generate high-quality paired views.}
    \label{fig:training}
\end{figure}

\section{Introduction}\label{sec:intro}
Whole slide imaging is an integral part of tissue diagnosis and laboratory medicine. Computational pathology can provide rapid tissue analysis of complex WSIs, such as cancer detection, subtyping, and grading. Modern computational methods have advanced beyond morphology-based diagnostics to ``omics" predictions, prognostication, and treatment response predictions from WSIs alone \cite{hollon2023artificial, coudray2018classification, chen2022pan}. Supervised multiple instance learning (MIL) methods, including ABMIL \cite{ilse2018attention}, CLAM \cite{lu2021data}, DSMIL \cite{li2021dual}, TransMIL \cite{shao2021transmil}, can achieve good performance on the above diagnostic tasks. Unfortunately, these methods rely on slide annotations to learn whole slide representations. WSI annotations are weak, sparse, incomplete, and expensive to obtain \cite{campanella2019clinical}. Moreover, WSIs are gigapixel-size and contain diverse morphologic and histopathologic features with extensive spatial heterogeneity. Weak slide labels may annotate only a small region within WSIs, demonstrating the limitation of relying on weak supervision to achieve high-quality and transferable whole slide representations.

To compensate for weak whole slide labels, self-supervised learning (SSL) has been increasingly used in computational pathology. The majority of previous SSL work in computational pathology has focused on region or patch learning. HiDisc, PLIP, and UNI are examples of WSI patch encoders that can be used to obtain visual features for downstream whole slide learning \cite{jiang2023hierarchical, chen2024towards, huang2023visual}. However, few previous studies have investigated learning whole slide representations using SSL \cite{lazard2023giga, chen2022scaling}. A general and unified framework for whole slide SSL would enable transferable whole slide feature learning and generalize to a wide range of downstream pathology tasks with minimal to no annotations required.

In this paper, we present Slide Pre-trained Transformers (SPT) for self-supervised whole slide representation learning. SPT treats gigapixel WSIs as a sequence of patch tokens and applies a domain-informed set of vision-language transformations, including splitting, cropping, and masking, to generate distinct views for self-supervised training. Over a range of patch encoders, SPT learns high-quality patch feature aggregation and whole slide representations compared to the state-of-the-art baselines. The main contributions are:

\begin{itemize}
    \item We introduce SPT, a general, flexible, and unified learning framework for WSIs at scale and benchmark on five computational pathology tasks.
    \item SPT outperforms previous state-of-the-art self-supervised and supervised methods for WSI representation learning. 
    \item SPT offers a consistent performance boost across a wide range of patch encoders.
\end{itemize} 

\section{Related Work}

\begin{figure*}[tb!]
    \centering
    \includegraphics[width=\textwidth]{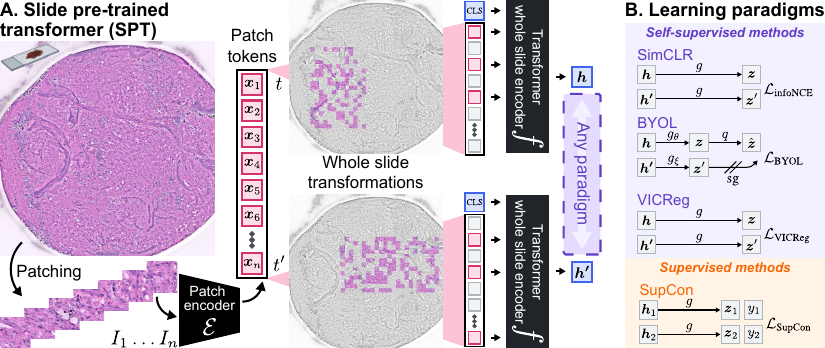}
    \caption{\textbf{SPT overview. }\textbf{A. }The SPT framework consists of a two-stage model architecture: 1) a pre-trained patch encoder $\mathcal{E}$ and 2) a transformer whole slide encoder $f$. WSIs are first divided into small patches, and the patch encoder extracts patch-level features. We then apply whole slide transformations to the patch tokens to create two views of the same WSI. The transformations combine splitting, cropping, and masking, which are informed by the structure and unique properties of WSIs. The transformed views are encoded by the transformer whole slide encoder, and the slide-level feature learning can use any paradigm. \textbf{B.} Example learning paradigms. In our experiments, we focus on three representative self-supervised paradigms, including SimCLR \cite{chen2020simple}, BYOL \cite{grill2020bootstrap}, and VICReg \cite{bardes2021vicreg}, and supervised contrastive learning \cite{khosla2020supervised}.}
    \label{fig:main.mech}
\end{figure*}

\subsection{Computational pathology}
Computational pathology combines whole slide imaging and computer vision methods for the analysis of biomedical tissues. Cancer diagnosis, prognostication, and response-to-treatment are some of the most common computer vision tasks within computational pathology. WSIs pose a unique computer vision challenge due to image sizes ranging up to 150K $\times$ 150K pixels. Data annotations are often limited to whole slide or patient-level labels \cite{campanella2019clinical}. Moreover, WSIs have a unique data structure compared to natural images, such as being non-object-centric and containing regional heterogeneity and visual feature redundancy. Despite these challenges, significant progress has been made over the last decade due to more accessible slide scanners and modern deep-learning methods. Cancer diagnosis \cite{campanella2019clinical, lu2021ai, lu2021data, hollon2020near}, molecular classification \cite{coudray2018classification, hollon2023artificial}, and prognostication \cite{chen2021multimodal} with WSIs represent how computational pathology can be used to improve diagnostic medicine and patient care. Whole slide representation learning for gigapixel WSI search and public WSI-natural language supervision represent future directions in computational pathology \cite{chen2022fast, huang2023visual}.

\subsection{Multiple instance learning}
Multiple instance learning (MIL) is a type of supervised learning such that labels are available only for bags of instances, rather than individual instances \cite{dietterich1997solving}. Learning to classify WSIs has generally been regarded as a MIL task, where each patch is an instance and the WSI is a bag of patches \cite{ilse2018attention}. There are two main components in MIL for WSI: the patch encoder and aggregator. Early works directly train the patch encoder in a weakly supervised fashion with WSI labels and generate a whole slide representation with mean pooling \cite{zhu2017wsisa,hou2016patch}. Many works have used ResNet or related architectures for patch encoders, often pre-trained on ImageNet \cite{yao2019deep,yao2020whole, campanella2019clinical,li2019patch, lu2021data, shao2021transmil}. Attention-based multiple instance learning (ABMIL) was first introduced in 2018 for WSI classification \cite{ilse2018attention}. Subsequently, clustering-constrained attention multiple instance learning (CLAM) was used for large-scale weakly supervised computational pathology. Subsequent MIL work has used variations of attention and more powerful transformer-based aggregation models, such as TransMIL \cite{li2019patch,yao2020whole,li2021dual,lu2021data,shao2021transmil,wang2022transformer,yu2023local,lin2023interventional}.

\subsection{Self-supervised representation learning}
Self-supervised learning (SSL) is a method of learning representations that does not require labels or annotations. SSL defines pretext learning tasks, such as instance discrimination or reconstruction, as training objectives and is typically evaluated by the performance of the learned representations on downstream tasks \cite{jaiswal2020survey}. SSL has shown success in natural images \cite{chen2020simple, grill2020bootstrap, bardes2021vicreg}, natural language processing \cite{mikolov2013efficient, devlin2018bert,radford2019language,brown2020language}, and multimodal learning \cite{huang2023visual, lu2023visual}. SSL has been used for patch encoding in WSIs. DSMIL \cite{li2021dual} applied SimCLR to learn patch representations. Most recently, PLIP \cite{huang2023visual} leverages histology images and comments posted to Twitter for self-supervised vision-language pretraining. UNI \cite{chen2024towards} built a general-purpose patch encoder for all types of H\&E patches and tasks training with self-supervised DINOv2 \cite{oquab2023dinov2} on a large-scale institutional dataset.

While there have been attempts at learning whole slide representations in an end-to-end manner with supervision \cite{hemati2021cnn}, or enhancing them with a supervised contrastive loss \cite{wang2022scl}, few works have focused on self-supervision beyond patches. HiDisc \cite{jiang2023hierarchical} uses a data hierarchy to learn patch representations that are invariant to slide and patient membership. HIPT \cite{chen2022scaling} uses a self-supervised patch encoder and an encoder for local regions, but weak supervision is used for learning WSI representations. Giga-SSL \cite{lazard2023giga} and SS-MIL \cite{tavolara2022contrastive} attempted to learn whole slide representations using self-supervision by masking patch tokens. Giga-SSL also emulates slide augmentations using pre-computed augmented patch representations. However, this strategy is memory intensive, does not span all combinations of possible random augmentations, and is infeasible for large-scale datasets. 

\section{Methods}
\subsection{The SPT framework}
WSIs are partitioned into smaller non-overlapping fields-of-view or patches. Let $\boldsymbol{I}=\left\{I_i\right\}_{i=1}^n$ be a WSI, where $n$ is the number of patches in a WSI, and each $I_i\in\mathbb{R}^{3\times H \times W}$, with coordinate $p_i \in\mathbb{Z}^2$. $\boldsymbol{p}=\left\{p_i\right\}_{i=1}^n$ are the coordinates of all patches in the WSI. SPT is a two-stage learning framework to learn WSI representations: 1) a pre-trained patch encoder $\mathcal{E}$; and 2) a transformer whole slide encoder $f$. The pre-trained patch encoder $\mathcal{E}$ can have the architecture of any visual feature extractor. The overall model architecture of SPT is illustrated in Figure \ref{fig:main.mech}.

In the SPT framework, the patch encoder $\mathcal{E}$ encodes each $I_i$ into a patch token, ${x}_i=\mathcal{E}(I_i)$. We represent all patch tokens in a WSI as $\boldsymbol{x} = \{x_i\}_{i=1}^n = \mathcal{E}(\boldsymbol{I})$. The whole-slide encoder $f$ serves as an aggregation function that learns a whole-slide representation using patch tokens and their corresponding coordinates: $h=f(\boldsymbol{x}, \boldsymbol{p})=f(\mathcal{E}(\boldsymbol{I}), \boldsymbol{p})$. $f$ can be any learned aggregation architecture, such as in ABMIL \cite{ilse2018attention} or a transformer \cite{shao2021transmil, vaswani2017attention}. Due to the scale of WSIs, it is infeasible to both train the patch encoder $\mathcal{E}$ and whole-slide encoder $f$ jointly in an end-to-end manner. Thus, we freeze the patch encoder $\mathcal{E}$ to allow for large mini-batch training of the whole-slide encoder $f$. \emph{The aim of the SPT framework is to learn high-quality whole slide representations with $f$ using self-supervision.}

Self-supervised methods share a common strategy: apply random transformations to a single data example to generate distinct views, called positive pairs. We designed our SPT framework to be compatible with existing SSL objectives, where a WSI undergoes transformations into different views. With a two-view SSL paradigm, $t(\boldsymbol{I})=[I_i \dots I_k]$, $t(\boldsymbol{p})=[p_i \dots p_k]$, $t'(\boldsymbol{I})=[I_j \dots I_\ell]$, and $t'(\boldsymbol{p})=[p_j \dots p_\ell]$, where $t, t'\in\mathcal{T}$ are randomly drawn from a set of transformations described in section \ref{sec:wsi.aug}. The patches and their corresponding coordinates from these transformed views are processed through $\mathcal{E}$ and $f$ to obtain the whole slide representations: 
\begin{equation*}
    h=f\left(\mathcal{E}(t(\boldsymbol{I})), t(\boldsymbol{p})\right),\quad h'=f\left(\mathcal{E}(t'(\boldsymbol{I})), t'(\boldsymbol{p})\right).
\end{equation*}
An SSL loss is used to self-supervise whole slide training:
\begin{equation*}
     \mathcal{L}_\text{SPT}\left(h,h'\right).
\end{equation*}
In our experiments, we selected representative methods from different SSL families \cite{balestriero2023cookbook}, including SimCLR \cite{chen2020simple} from contrastive learning, BYOL \cite{grill2020bootstrap} from self-distillation, and VICReg \cite{bardes2021vicreg} from canonical component analysis. We denote the self-supervised SPT as \emph{\textbf{ssSPT}}.

\paragraph{SPT with supervision.} SPT can be adapted to fully supervised training using weak slide- or patient-level labels, by applying a supervised contrastive loss. Positive pairs for supervised contrastive learning are defined by class labels \cite{khosla2020supervised}. We denote the supervised variant of SPT as \emph{\textbf{suSPT}}.

\subsection{SPT transformation}\label{sec:wsi.aug}

\begin{figure}[htb!]
    \centering
    \includegraphics[width=\columnwidth]{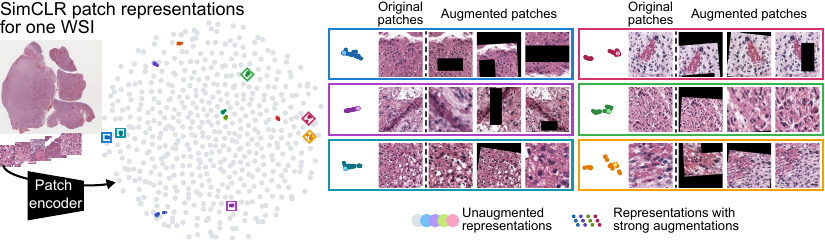}
    \caption{\textbf{Limited effect of pixel-level patch augmentations.} We qualitatively evaluate the effect of pixel-level augmentation on the patch representations by visualizing the \emph{t}SNE plot of SimCLR pre-trained patch representations sampled from a single WSI. We observe that strong augmentations at the pixel level have a minimal effect on the patch embeddings. The invariant behavior of the patch encoder is explicitly enforced by the SimCLR pretext task.}
    \label{fig:main.patch_aug}
\end{figure}

While previous work focused on pixel space transformations~\cite{lazard2023giga}, we hypothesized that these augmentations at patch level are insufficient to generate high-quality views for whole slide SSL. A patch encoder trained with instance discrimination, for example, should be invariant to pixel-space augmentations. 
Thus, pixel space augmentations, such as color jittering, should have minimal effect on the representation space and resulting similar embeddings, as illustrated with SimCLR in Figure \ref{fig:main.patch_aug}, because it is explicitly enforced as the pretext task. 
We opt to bypass patch augmentations altogether, thereby reducing SPT memory and compute burden, and focusing on transformations at WSI level.

\begin{figure*}[tb!]
    \centering
    \includegraphics[width=\textwidth]{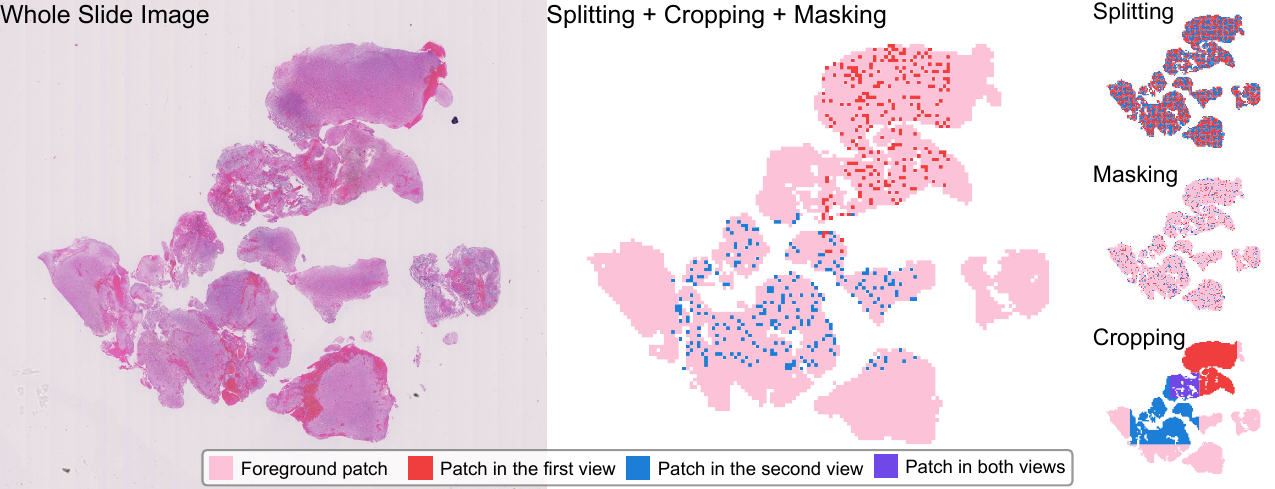}
    \caption{\textbf{SPT transformation strategy.} SPT combines splitting, cropping, and masking to generate views, and they are motivated by the size, region diversity, and information redundancy of WSIs. Splitting partitions patches into mutually exclusive sets decreases mutual information between views; cropping can generate spatially diverse views covering different regions on the WSI; masking reduces redundant visual features and improves training efficiency. The combination of these transformations can create optimal positive pairs for whole slide representation learning.}
    \label{fig:main.aug}
\end{figure*}

\paragraph{Transformation strategy.} WSIs are divided into a sequence of patch tokens. The SPT transformation strategy, as shown in Figure \ref{fig:main.aug}, is inspired by both vision and language modeling, and was selected to address the domain-specific properties of WSI:

\begin{itemize}
     \item \textbf{Splitting} \textit{to decrease mutual information between views.} Reducing mutual information between SSL views, while keeping task-relevant information intact, improves downstream performance~\cite{tian2020makes, Shwartz_Ziv2024-jl}. The splitting transformation randomly partitions patch tokens into disjoint sets, ensuring mutually exclusive set membership for each token. Enforcing mutual exclusion reduces mutual information between SSL views.
     
     \item \textbf{Cropping} \textit{to capture regional heterogeneity}. WSIs contain regional differences in histologic features. Similar to multi-cropping in visual SSL \cite{chen2020simple, caron2020unsupervised, caron2021emerging}, cropping generates spatially diverse views of WSI with variable sizes and features.
     
     \item \textbf{Masking} \textit{to reduce redundant visual features}. Due to their gigapixel size, WSIs can contain large regions of similar histologic features and tissue phenotypes, causing redundancy. Masking is an effective transformation for vision and language modeling \cite{chen2022scaling, devlin2018bert, he2022masked}. For each view, the masking transformation samples a subset of patch tokens without replacement. Additionally, masking increases training efficiency by reducing sequence lengths.
\end{itemize}

While splitting, cropping and masking can be used individually, we apply them jointly to generate versatile and high-quality views for SPT training. 

\subsection{SPT Implementation} \label{sec:main.methods.impl}
End-to-end whole slide learning with large batch sizes is infeasible due to the gigapixel scale. To enable efficient training, all unaugmented patch tokens were computed on a frozen patch encoder, and SPT transformations were then applied to patch tokens. Each whole slide was represented as an embedding $\boldsymbol{x}\in\mathbb{R}^{n\times d}$ and a coordinate $\boldsymbol{p}\in\mathbb{Z}^{n\times 2}$. Splitting and masking were implemented as row-wise partitioning and dropout on $\boldsymbol{x},\  \boldsymbol{p}$, respectively. Cropping was implemented as coordinate filtering on $\boldsymbol{p}$. A detailed description of our implementation with pseudocode is in Appendix \ref{sec:supp.methods}.

\section{Experiments} \label{sec:main.exp}

\subsection{Benchmarks} \label{sec:main.exp.data}
We evaluated SPT visual representations on five benchmarks across three clinical tasks, including two different imaging modalities. Additional dataset information and breakdown are in Appendix \ref{sec:supp.data}.

\paragraph{SRH CNS benchmark.}
Stimulated Raman Histology (SRH) is a novel optical microscopy method that enables fast imaging of unprocessed tissues \cite{Freudiger2008-gj,orringer2017rapid}. The benchmark includes six central nervous system (CNS) tumors and normal brains, with 2035 and 925 WSIs for training and evaluation, respectively. These data were collected at University of Michigan, following the imaging protocol in \cite{jiang2022opensrh}, and were labeled by board-certified pathologists. The study has been approved by the Institutional Review Board (HUM00083059), with informed consent from each patient.

\paragraph{H\&E glioma molecular classification benchmark.} 
We also evaluated SPT using publicly available diffuse glioma H\&E stained WSIs from the Cancer Genome Atlas (TCGA) and Digital Brain Tumour Atlas (DBTA) \cite{roetzer2022digital}. We focused on the classification of three molecular subgroups, as defined by the World Health Organization \cite{louis20212021}. Molecular classification is a challenging computer vision task because the diagnoses are made via molecular testing (such as genetic sequencing), and are not possible for expert pathologists using H\&E images alone. Our glioma dataset is comprised of 2309 training slides and 341 slides from the TCGA dataset set for evaluation.

\paragraph{TCGA BRCA, TCGA NSCLC, and TCGA RCC benchmarks.}
We further evaluated SPT using three widely used TCGA H\&E classification benchmarks: 1) invasive breast carcinoma (BRCA) subtyping, 2) non-small cell lung carcinoma (NSCLC) subtyping, and 3) renal cell carcinoma (RCC) subtyping. For these benchmarks, we followed the well-established study design of \cite{chen2022scaling}.

\subsection{Implementation details} \label{sec:main.exp.imp}
We trained in-domain ResNet-34 \cite{he2016deep} patch encoders to extract features for all patches in each slide. These patch encoders were trained with SimCLR \cite{chen2020simple} and HiDisc \cite{jiang2023hierarchical}. Additional off-the-shelf and foundational patch encoders are described in Figure \ref{tab:results.patch_encoders}. SPT slide encoders are six-layer transformers \cite{vaswani2017attention} with four heads, and learnable Fourier positional embeddings \cite{li2021learnable} (visualized in Appendix \ref{sec:supp.methods.position}). We used a two-layer projection head for SimCLR and VICReg, and one-layer projection and prediction heads for BYOL. We used AdamW optimizers and cosine decay schedulers after warm-up in the initial 10\% of the iterations. The learning rate was adjusted between 10\textsuperscript{-3} and 10\textsuperscript{-7} to accommodate the training dynamics of different SSL methods and tasks. With our SPT slide transformation strategy, we adjusted cropping and masking sizes for each experiment and utilized up to 64 patches per slide for each view. We trained ssSPT and suSPT experiments up to 800 and 100 epochs, respectively, with an effective batch size of up to 1024 WSIs. All models were trained with mixed-precision on a NVIDIA A40 GPU, taking up to 8 hours.

\subsection{Evaluation Protocol}\label{sec:main.exp.eval}

We benchmarked SPT using standard linear evaluation protocols. Since linear classifiers are sensitive to hyperparameters \cite{caron2021emerging}, we also employed \emph{k} nearest neighbor (\emph{k}NN) for direct evaluation. H\&E glioma molecular evaluation used only WSI from TCGA. For SRH and H\&E Glioma, experiments were repeated with three random seeds, and for TCGA BRCA, NSCLC, and RCC, we used 10-fold cross-validation for error bars, following the protocols in previous work \cite{chen2022scaling, lazard2023giga}. For the baselines, original embeddings or logits were used when available. We used mean class accuracy (MCA), F1 scores, and area under the receiver operating characteristic (AUC) to evaluate all benchmarks.

\section{Results} \label{sec:main.results}
We first benchmarked ssSPT performance with self-supervised WSI learning strategies in section \ref{sec:main.results.ssspt}. We compared ssSPT and suSPT with supervised WSI MIL strategies in section \ref{sec:main.results.suspt}. We then evaluated the ability of SPT to generalize across a wide range of patch encoders in section \ref{sec:main.results.patchenc}. Next, we showed that SPT improves MIL results with SOTA foundation model patch encoders in section \ref{sec:main.results.uni}. Finally, we visualized SPT self-attention heatmaps in section \ref{sec:main.results.heatmap}. Full results with error bars, ablation studies on the \textbf{SPT transformations and parameters,} and additional visualizations are in Appendix \ref{app:ext:result}.

\subsection{ssSPT learns high-quality whole slide representations} \label{sec:main.results.ssspt}

We benchmarked ssSPT with self-supervised baselines using in-domain patch encoders. WSI features were evaluated directly using a \emph{k}NN classifier. As shown in Table \ref{tab:main.results.ssspt.knn}, ssSPT surpasses all baselines across all self-supervised objectives in all metrics, except for AUC on the highly imbalanced BRCA benchmark. ssSPT outperforms all existing self-supervised methods by a large margin, outperforming the previous best self-supervised method with a 10 and 5 points increase in MCA on the SRH CNS and H\&E glioma benchmarks, respectively.

\newcommand\dsetheaders{& \multicolumn{3}{c}{SRH CNS}
                      & \multicolumn{3}{c}{H\&E Glioma}
                      & \multicolumn{3}{c}{TCGA BRCA}
                      & \multicolumn{3}{c}{TCGA NSCLC}
                      & \multicolumn{3}{c}{TCGA RCC}}
\newcommand\metricheaders{MCA & F1 & AUC}

\begin{table*}[h!]
\setlength{\tabcolsep}{2pt}
\centering{\resizebox{\textwidth}{!}{

\begin{tabular}{cccccccccccccccc}\toprule
\dsetheaders\\
\cmidrule(lr){2-4}\cmidrule(lr){5-7}\cmidrule(lr){8-10}\cmidrule(lr){11-13}\cmidrule(lr){14-16}
 &    \metricheaders  &     \metricheaders  &     \metricheaders  &     \metricheaders  &     \metricheaders\\\midrule
 \multicolumn{16}{c}{\cellcolor{violet2}\emph{Self-supervised methods}}\\\addlinespace
Pooling      &     72.5  &     73.2  &     94.8  &     68.3  &     67.9  &     87.9  &     58.1  &     27.4  &     74.7  &     75.1  &     76.5  &     85.3  &     84.2  &     86.3  &     96.6\\
HIPT  &     -     &     -     &     -     &     -     &     -     &     -     &     63.2  &     39.9  &     77.5  &     80.6  &     80.1  &     88.9  &     88.4  &     89.2  &     97.4\\
Giga-SSL     &     71.2  &     72.7  &     94.2  &     71.6  &     71.5  &     89.1  &     70.4  &     53.9  &     \tb{85.4} & 84.3  &     84.9  &     \tb{92.3}  &     90.0  &     89.3  &     97.5\\
ssSPT (Ours) & \tb{82.3} & \tb{82.3} & \tb{94.7} & \tb{76.5} & \tb{76.1} & \tb{90.9} & \tb{72.7} & \tb{58.3} & 80.4  &     \tb{86.1} & \tb{86.1} & \tb{92.3} & \tb{91.7} & \tb{90.5} & \tb{98.3}\\\bottomrule
\end{tabular}}}

\caption{\textbf{Self-supervised benchmarks.} We use $k$NN classifier to evaluate ssSPT and baselines. We report the best performing SSL objective for ssSPT, with additional SSL objectives in Appendix \ref{app:ext:result} Table \ref{tab:supp.results.ssspt.ablation.algo}. Mean values are reported here and standard deviations are in Appendix \ref{app:ext:result} Table \ref{tab:supp.results.ssspt.knn}.}
\label{tab:main.results.ssspt.knn}
\end{table*}

\subsection{SPT outperforms previous fully supervised methods}\label{sec:main.results.suspt}

We benchmarked ssSPT and suSPT with linear evaluation using in-domain patch encoders. As shown in Table \ref{tab:main.results.suspt.linear}, SPT outperforms previous fully supervised methods across all tasks. Remarkably, ssSPT outperforms or matches previous \emph{fully supervised} methods on MCA in SRH CNS (+1.0), H\&E glioma (match), TCGA BRCA (+2.2), and TCGA RCC (+1.4) benchmarks. suSPT outperforms existing MIL methods across all five benchmarks on nearly all metrics. In comparison with the best-performing baselines on MCA of these five benchmarks, suSPT achieves a performance increase of 1.2, 0.7, 3.4, 2.1, and 0.3 points, respectively. Thus, SPT provides a performance increase for both self-supervised and supervised learning, generalizing to different tissue types and diagnostic tasks.

\begin{table*}[htb!]
\setlength{\tabcolsep}{2pt}
\centering{\resizebox{\textwidth}{!}{

\begin{tabular}{cccccccccccccccc}\toprule
\dsetheaders\\
\cmidrule(lr){2-4}\cmidrule(lr){5-7}\cmidrule(lr){8-10}\cmidrule(lr){11-13}\cmidrule(lr){14-16}
&\metricheaders & \metricheaders & \metricheaders & \metricheaders & \metricheaders\\\midrule
\multicolumn{16}{c}{\cellcolor{violet2}\emph{Self-supervised methods}}\\\addlinespace
Giga-SSL      &     78.6   &     75.6   &     96.1   &     76.5   &     73.9      &     90.7   &     81.7   &     62.6   & \tb{91.2}  &     87.5   &     87.9   &     93.7   &     89.9   &     87.8   &     97.7 \\
ssSPT (Ours) & \tb{85.4}  & \tb{83.2} & \tb{97.6} & \tb{80.0} & \tb{78.0} & \tb{91.9} & \tb{83.0} & \tb{67.4}  &     89.4   & \tb{88.0} & \tb{88.0} & \tb{94.8} & \tb{93.4} & \tb{91.1} & \tb{98.7} \\\addlinespace
\multicolumn{16}{c}{\cellcolor{orange2}\emph{Supervised methods}}\\\addlinespace
ABMIL        &     79.8   &     78.9   &     96.0   &     75.0   &     74.5      &     90.2   &     78.6   &     62.8   &     86.1   &     86.8   &     86.7   &     93.7   &     89.9   &     89.4   &     97.8 \\
CLAM         &     83.5   &     82.9   &     96.7   &     78.9   &     77.7      &     88.5   &     76.8   &     59.3   &     85.8   &     84.3   &     83.6   &     92.8   &     85.6   &     83.6   &     97.3 \\
DSMIL        &     -      &     -      &     -      &     -      &     -         &     -      &     77.0   &     58.0   &     83.8   &     84.9   &     84.0   &     92.0   &     90.1   &     87.1   &     97.1 \\
TransMIL     &     84.4   &     84.0   &     84.0   &     80.0   &     79.4      &     91.9   &     80.8   &     69.4   &     90.0   &     87.4   &     87.5   &     95.3   &     87.4   &     87.5   &     95.3 \\
HIPT         &     -      &     -      &     -      &     -      &     -         &     -      &     75.8   &     60.2   &     87.4   &     88.2   &     88.2   &     95.2   &     92.0   & \tb{92.1}  &     98.0 \\
suSPT (Ours) & \tb{85.6} & \tb{84.4} & \tb{97.2} & \tb{80.7} & \tb{80.3} & \tb{93.7} & \tb{84.2} & \tb{70.4} & \tb{90.5} & \tb{90.3} & \tb{90.4} & \tb{95.7} & \tb{92.3} &     91.0 & \tb{98.6} \\\bottomrule
\end{tabular}}}
\caption{\textbf{SPT benchmarks.} We report linear evaluation results on five histology benchmarks. We report the best performing SSL objective for ssSPT, with additional SSL objectives in Appendix \ref{app:ext:result} Table \ref{tab:supp.results.ssspt.ablation.algo}. Mean values are reported here and standard deviations are in  Appendix \ref{app:ext:result} Table \ref{tab:supp.results.suspt}.}
\label{tab:main.results.suspt.linear}
\end{table*}

\subsection{SPT offers performance boost across a range of patch encoders} \label{sec:main.results.patchenc}

\begin{figure}[h]
     \centering
     \begin{subfigure}[c]{0.50\textwidth}
        \centering{\resizebox{\textwidth}{!}{
        \setlength{\tabcolsep}{2pt}
        \begin{tabular}{cccc}\toprule
        \begin{tabular}{@{}c@{}}Patch\\encoder\end{tabular}   &
        Architecture & 
        \begin{tabular}{@{}c@{}}Training\\dataset\end{tabular} & \\\midrule
        ImageNet \cite{he2016deep}           & ResNet34    & ImageNet                                 \\\midrule
        HIPT \cite{chen2022scaling}          & ViT-S/16    & All of TCGA                              \\
        SimCLR \cite{chen2020simple}         & ResNet34    & In-domain (see \ref{sec:main.exp.imp})  \\
        HiDisc \cite{jiang2023hierarchical}  & ResNet34    & In-domain (see \ref{sec:main.exp.imp})  \\\midrule
        PLIP \cite{huang2023visual}          & ViT-B/32    & OpenPath                                 \\
        UNI \cite{chen2024towards}           & ViT-L/16    & Mass-100K + GTEx                         \\\bottomrule
        \end{tabular}
        }}
        \caption{\textbf{Patch encoders.} ImageNet is an off-the-shelf model trained on OOD data. HIPT is near-domain for H\&E glioma. SimCLR and HiDisc patch encoders are trained with in-domain data. PLIP and UNI are SOTA foundational patch encoders.}
        \label{tab:results.patch_encoders}
     \end{subfigure}
     \hfill
     \begin{subfigure}[c]{0.48\textwidth}
        \includegraphics[width=\textwidth]{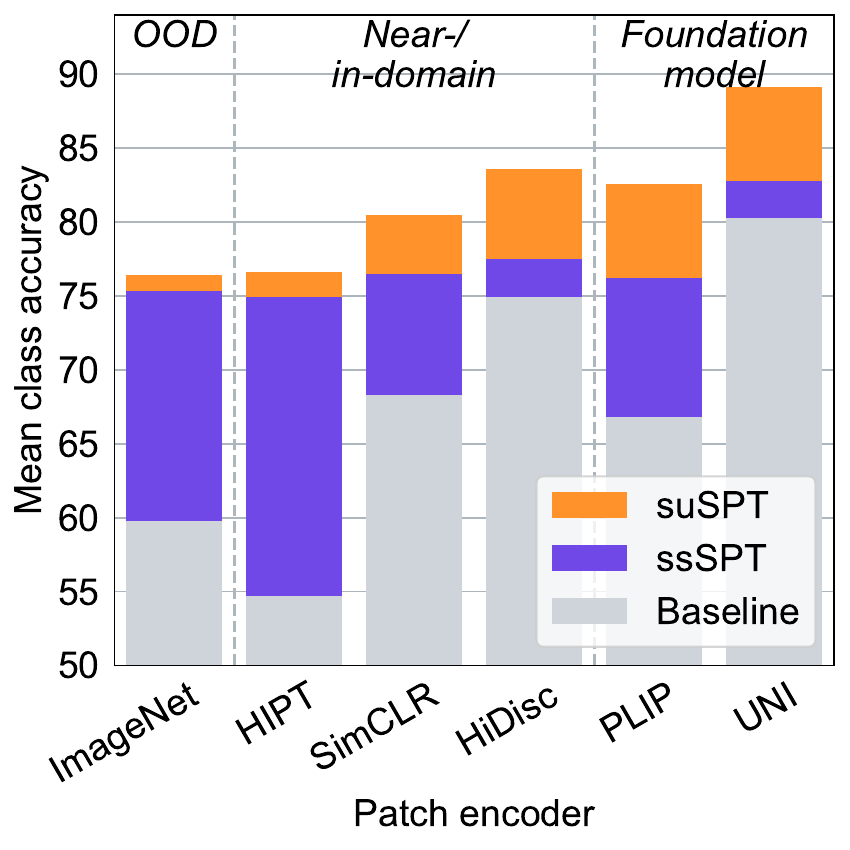}
        \caption{\textbf{SPT results on H\&E Glioma.}}
        \label{fig:main.results.patch_enc}
     \end{subfigure}
     \caption{\textbf{SPT benchmarks with different patch encoders.} ssSPT and suSPT offer performance boosts with a wide range of patch encoders. ssSPT approaches supervised performance upperbound. Additional metrics with error bars are in  Appendix \ref{app:ext:result} Table \ref{tab:supp.results.patch_encoders}.}\label{fig:results.patch_encoders_comp}
\end{figure}

We explored the generalizability of SPT to different patch encoders using the more challenging H\&E Glioma benchmark. As described in Table \ref{tab:results.patch_encoders}, we tested ImageNet, HIPT \cite{chen2022scaling}, PLIP \cite{huang2023visual}, and UNI \cite{chen2024towards}, in addition to in-domain patch encoders trained with SimCLR and HiDisc. The ImageNet patch encoder is off-the-shelf and out-of-distribution (OOD). HIPT \cite{chen2022scaling} is near-domain since it was trained with H\&E WSIs from TCGA, including multiple organ systems and institutions. PLIP \cite{huang2023visual} and UNI \cite{chen2024towards} are visual foundational models for histology, trained with OpenPath and large-scale institutional datasets, respectively. SPT performance is shown in Figure \ref{fig:main.results.patch_enc}, with additional results for SRH CNS benchmark in Appendix \ref{app:ext:result} Figure \ref{fig:supp.results.patch_enc.srh}, and extended metrics with error bars are in Table \ref{tab:supp.results.patch_encoders}. 

For all patch encoders, SPT training significantly improves whole slide representations over pooling baselines. This improvement is the most significant for ImageNet and HIPT patch encoders. As expected, the performance boost is smaller for in-domain patch encoders, especially HiDisc, where slide discrimination was learned during patch training. 
SPT also achieves a large performance boost using PLIP patch encoders, as it bridges the domain gap between our benchmark and the pre-training dataset  OpenPath which is collected on Twitter. As for the state-of-the-art UNI patch encoder, we still observed a large performance boost achieving MCA near 90 points for suSPT. Overall, these results demonstrate that SPT can enhance whole-slide representation learning using encoders from different domains, thereby reducing the reliance on a specialized in-domain patch encoder. \emph{SPT training time ranges from 6-8 GPU hours, while the patch encoder is computationally intensive and usually took over 80 GPU hours to train.} 

\subsection{SPT improves state-of-the-art MIL results} \label{sec:main.results.uni}

Most recently, foundation models have been developed for computational pathology, promising a general approach for WSI diagnostic tasks. UNI is a foundational patch encoder that achieves state-of-the-art results \cite{chen2024towards}. The UNI authors found that ``ABMIL with UNI features outperforms many sophisticated MIL architectures'' \cite{chen2024towards}. We benchmarked suSPT using UNI patch features, in comparison to ABMIL in Table \ref{tab:main.results.suspt.uni}. suSPT outperforms ABMIL in all benchmarks with UNI features, improving upon the previous state-of-the-art performance on these tasks. This showcases that SPT is complementary to the innovative medical foundation models for learning patch representations. 

\begin{table*}[h!]
\setlength{\tabcolsep}{2pt}
\centering{\resizebox{\textwidth}{!}{
\begin{tabular}{cccccccccccccccc}\toprule
\dsetheaders\\
\cmidrule(lr){2-4}\cmidrule(lr){5-7}\cmidrule(lr){8-10}\cmidrule(lr){11-13}\cmidrule(lr){14-16}
 &    \metricheaders  &      \metricheaders  &      \metricheaders  &      \metricheaders  &      \metricheaders\\\midrule
UNI + ABMIL   &     86.3  &     86.3  &     96.3  &     86.8  &     85.2  &     95.5  &     88.6  &     79.1  &     88.8  &     93.6  &     90.6  &     93.4  &     95.2  &     93.3  &     98.5\\
UNI + suSPT   & \bf{86.4} & \bf{86.8} & \bf{98.4} & \bf{89.1} & \bf{88.7} & \bf{97.3} & \bf{89.6} & \bf{82.2} & \bf{94.7} & \bf{95.5} & \bf{95.6} & \bf{97.6} & \bf{96.2} & \bf{94.9} & \bf{98.9}\\\bottomrule
\end{tabular}}}
\caption{\textbf{suSPT improves state-of-the-art MIL results with UNI patch features.} Mean values are reported here, and standard deviations are in Appendix \ref{app:ext:result} Table \ref{tab:supp.results.suspt.uni}.} 
\label{tab:main.results.suspt.uni} 
\end{table*}

\subsection{Self-attention visualizations reveal tissue phenotypes in full gigapixel WSIs} \label{sec:main.results.heatmap}

\begin{figure*}[h!]
    \centering
    \includegraphics[width=\textwidth]{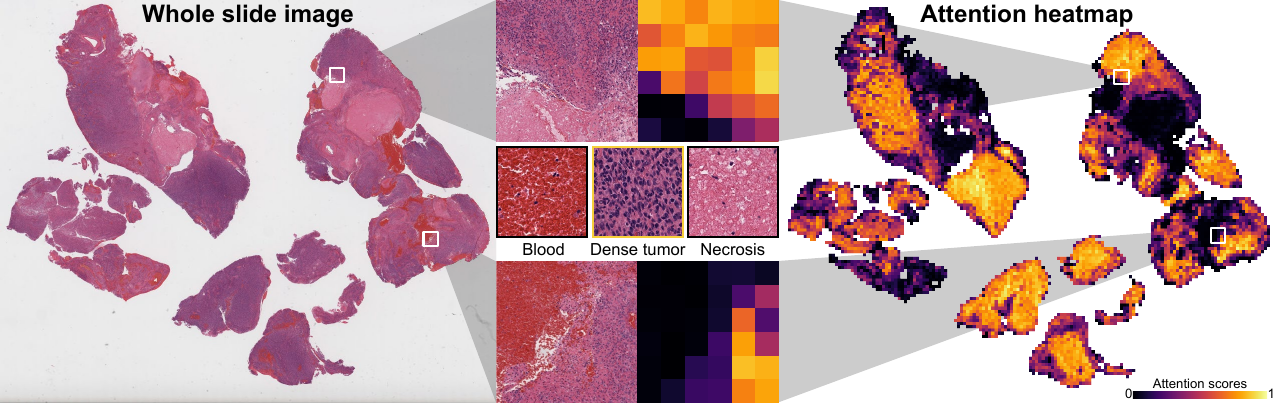}
    \caption{\textbf{Attention heatmap visualization.} Self-attention visualization demonstrates ssSPT's ability to distinguish between different tissue phenotypes such as blood, dense tumor, and necrosis. To our knowledge, these are the first self-supervised transformer-based attention maps on full gigapixel WSIs, instead of localized regions.}
    \label{fig:main.results.heatmap}
\end{figure*}

Finally, we adopted the strategy in \cite{caron2021emerging} and evaluated ssSPT qualitatively by visualizing self-attention on the full WSI. As illustrated in Figure \ref{fig:main.results.heatmap}, self-attention maps on H\&E WSI can distinguish different tissue phenotypes such as blood, dense tumor, and necrosis. The CLS token attends to dense tumor regions instead of non-diagnostic regions such as blood and necrosis for slide representations. This observation is consistent with existing literature on self-supervised vision transformers, where attention maps can serve as unsupervised segmentation \cite{caron2021emerging, chen2022scaling}. To our knowledge, \emph{our work is the first to generate self-supervised transformer-based attention maps on full gigapixel WSIs, instead of localized regions,} showing the strong representation capacity, flexibility, and efficiency of ssSPT. More H\&E and SRH attention map visualizations are in Appendix \ref{sec:supp.results.attention}.

\section{Conclusion}\label{sec:conclu}
We present \emph{Slide Pre-trained Transformers (SPT),} a self-supervised approach for whole slide representation learning, in a general, versatile, and lightweight framework. SPT uses domain-informed, vision-language transformations for high-quality WSI view generation and self-supervised learning. We demonstrated that SPT achieves superior performance on five computational pathology benchmarks across three clinical tasks, including histopathologic diagnosis, cancer subtyping, and molecular genetic prediction. To our knowledge, this is the first systematic study of whole slide representation learning evaluating the role of SSL objectives, transformation strategies, and patch encoders. 

\paragraph{Limitations.} Our experimentation was limited to two biomedical microscopy modalities and three major self-supervised learning paradigms. Future work includes validating SPT on additional datasets, including WSIs from different microscopy modalities, such as fluorescent microscopy, as well as different SSL paradigms such as generative SSL. Another limitation of SPT is hyperparameter tuning for different configurations of SPT transformations, which are domain and dataset-specific. The future direction of our work also includes efficient end-to-end training to unify the training of patch and slide encoders.

\paragraph{Broader impact.}
This paper aims to advance the field of whole slide representation learning and computational pathology. Our work builds toward an automated diagnostic support system for biomedical microscopy and histopathology, paving the way for more accurate diagnoses and personalized treatment recommendations. Moreover, we strive to develop self-supervised learning frameworks in order to promote and maintain data privacy and ethical use of patient data within healthcare AI systems. Additionally, SSL is essential for developing medical foundation models. We hope that SPT can serve as a potential foundation model training strategy within computational pathology. While our experiments are in medical applications, there is a potential for impact on other similar gigapixel image modalities, such as geospatial and astronomical images.

\section*{Acknowledgements and Competing Interests}
We would like to thank Karen Eddy and Lin Wang for their administrative support and data collection efforts.

This work was supported, in part, by the National Institutes of Health (NIH) grants F31NS135973 (C.J.), T32GM141746 (C.J.), and K12NS080223 (T.H.). This work was also supported, in part, by the Chan Zuckerberg Foundation (CZI) Advancing Imaging Through Collaborative Project grant (T.H.), the Cook Family Brain Tumor Research Fund (T.H.), the Mark Trauner Brain Research Fund (T.H.), the Zenkel Family Foundation (T.H.), Ian’s Friends Foundation (T.H.) and the UM Precision Health Investigators Awards grant program (T.H.).

T.H. is a shareholder of Invenio Imaging, Inc., a company developing SRH microscopy systems.

\bibliographystyle{unsrt}
\bibliography{spt}

\clearpage
\appendix 

\section{Data Details} \label{sec:supp.data}
The five benchmarks used in our experimentation are described in Table \ref{tab:data_overview}.

\begin{table*}[h!]
    \centering{
    \begin{tabular}{ccccc}\toprule
        Benchmark   & \# clases & Tissue organ & Imaging modality & Task \\\midrule
        SRH CNS     & 7         & Brain        & SRH              & Histological classification\\
        H\&E Glioma & 3         & Brain        & H\&E stained     & Molecular classification\\
        TCGA BRCA   & 2         & Breast       & H\&E stained     & Tumor subtyping\\
        TCGA NSCLC  & 2         & Lung         & H\&E stained     & Tumor subtyping\\
        TCGA RCC    & 3         & Kidney       & H\&E stained     & Tumor subtyping\\\bottomrule
    \end{tabular}}
    \caption{\textbf{Benchmark overview.} Our benchmarks span four different organs and two different microscopy modalities.}
    \label{tab:data_overview}
\end{table*}

\subsection{SRH CNS benchmark} 
Our SRH benchmark includes common brain tumors and normal brain tissues. Our data collection and processing follows the protocol in \cite{jiang2022opensrh}, where each specimen is scanned with a commercially available stimulated Raman histology microscope manufactured by Invenio Imaging. Each specimen is diagnosed by a board-certified neuropathologist. The number of slides in each class is in Table \ref{tab:srh_stats}. 

\begin{table*}[h!]
    \centering{
    \begin{tabular}{ccc}\toprule
        Class & Training & Validation \\\midrule
        HGG            & 407        & 132\\
        LGG            & 210        & 107\\
        Mening.        & 434        & 204\\
        Metast.        & 236        & 114\\
        Pit.           & 448        & 194\\
        Schwan.        & 47         & 22\\
        Normal         & 253        & 152\\
        Unlabeled      & 2919       & - \\\bottomrule
    \end{tabular}}
    \caption{\textbf{SRH dataset breakdown.} Number of slides in the training and validation set for supervised training and evaluation. HGG, high grade glioma, LGG, low grade glioma, mening., meningioma, metast., metastasis, pit., pituitary adenoma, schwan., schwannoma, normal, normal brain tissue.}
    \label{tab:srh_stats}
\end{table*}

\subsection{H\&E Glioma benchmark} 
Our H\&E benchmark includes glioma specimens from the Cancer Genome Atlas Program (TCGA) and the Digital Brain Tumour Atlas (DBTA) \cite{roetzer2022digital}. These publicly available datasets feature H\&E images collected in both the United States and Europe. Slides are divided into three classes based on molecular labels as defined by the World Health Organization \cite{louis20212021}, using IDH and 1p/19q co-deletion status included in both datasets.  At training time, both DBTA and TCGA training set is used, and a separate, held-out validation set from the TCGA dataset is used for benchmarking. A detailed number of slides per class is in Table \ref{tab:he_stats}. Our H\&E data processing pipeline follows \cite{jiang2023hierarchical}. Each WSI is divided into 300 $\times$ 300 patches, and blank / background regions are excluded. All patches used in training are stain-normalized using the Macenko algorithm \cite{macenko2009method}.

\begin{table*}[h!]
    \centering{
    \begin{tabular}{ccccc}\toprule
        \multirow{2}{*}{Class} & \multirow{2}{*}{Molecular label} & \multicolumn{2}{c}{Training} & Validation\\
                           &                                 & DBTA       & TCGA       & TCGA \\\midrule
        Oligodendroglioma  &  IDH mutant, 1p/19q co-deleted  & 176        & 265        & 62\\
        Astrocytoma        &  IDH mutant                     & 157        & 360        & 88\\
        Glioblastoma       &  IDH wildtype                   & 619        & 732        & 191\\\bottomrule
    \end{tabular}}
    \caption{\textbf{H\&E glioma dataset breakdown.} Number of slides from both datasets.}
    \label{tab:he_stats}
\end{table*}

\section{Methods and Implementation} \label{sec:supp.methods}
\subsection{SPT transformation details}

As described in section \ref{sec:main.methods.impl}, we implement our SPT transformations with pre-computed embeddings and patch coordinates. Recall that each whole slide is represented as a $(\boldsymbol{x}, \boldsymbol{p})$ tuple, where $\boldsymbol{x}\in\mathbb{R}^{n\times d}$ is the embedding tensor and $\boldsymbol{p}\in\mathbb{Z}^{n\times 2}$ is a corresponding coordinate tensor for each patch in the WSI. Each transformation is implemented by indexing into the rows (corresponding to patches/tokens) of $\boldsymbol{x}$ and $\boldsymbol{p}$. A PyTorch style pseudocode is in Algorithm \ref{alg:aug}.

\paragraph{Splitting.} Splitting has one hyperparameter: the ratio of tokens between two views. The ratio specifies the fraction of tokens in the first view and is used to compute the number of tokens in each view. Tokens are split randomly between views accordingly, and two disjoint views are returned.

\paragraph{Cropping.} Cropping has two sets of hyperparameters: cropping area range, and cropping aspect ratio range. The cropping area and cropping aspect ratio are randomly chosen from their respective ranges.  A random token in the view with coordinates $(r_A, c_A)$ is selected to be an anchor as the crop center of the view. A coordinate range $(r_0, c_0), (r_1, c_1)$ is computed based on the area and aspect ratio:
\begin{align*}
    H &= \sqrt{\text{Area}/\text{aspect}}\\
    W &= H \cdot \text{aspect}\\
    r_0 &= r_A - H / 2; & r_1 &= r_A + H / 2\\
    c_0 &= c_A - W / 2; & c_1 &= c_A + W / 2,
\end{align*}
where $H, W$ are the height and width of the crop. Tokens in this coordinate range are included in the transformed view.

\paragraph{Masking.} Masking has two hyperparameters: masking ratio range and max masking token limit. The masking ratio is randomly drawn from the masking ratio range, and it is used to compute the number of tokens $m$ to keep in the augmented view. Max masking token limit is a cap of the number of tokens in each view, i.e. $m = \min(m, \text{max\_token\_lim})$. This parameter may be omitted (set to $\infty$) when training on datasets with smaller WSIs. Finally, we randomly select $m$ tokens to remain in the augmented view.

\begin{algorithm}[p!]
\caption{SPT transformations in PyTorch style}
\label{alg:aug}
\definecolor{codeblue}{rgb}{0.0,0.5,0.0}
\definecolor{codekw}{rgb}{0.85, 0.18, 0.50}
\lstset{
  basicstyle=\fontsize{8pt}{8pt}\ttfamily\selectfont,
  columns=fullflexible,
  breaklines=true,
  captionpos=b,
  commentstyle=\color{codeblue},
  keywordstyle=\color{codekw},
}
\begin{lstlisting}[language=python]
# hyperparameters:
# - split_ratio: fraction of token in first view
# - (area_min, area_max): cropping area range
# - (aspect_min, aspect_max): cropping aspect ratio range
# - (mask_min, mask_max): masking ratio range
# - max_token_lim: max masking token limit


@torch.no_grad()
def transform(x, p):
    # inputs:
    # - x: patch embeddings (n d)
    # - p: patch coordinates (n 2)
    # output:
    # - (x1, p1): patch embs and coords for the first view
    # - (x2, p2): patch embs and coords for the second view
    
    (x1, p1), (x2, p2) = split(x, p)
    (x1, p1) = mask(crop(x1, p1))
    (x2, p2) = mask(crop(x2, p2))
    return (x1, p1), (x2, p2)

    
def split(x, p):
    # randomly permute index of all tokens
    rand_idx = torch.randperm(len(p))
    
    # determine index for each view
    view1_sz = int(split_ratio * len(p))
    idx1, idx2 = rand_idx[:view1_sz], rand_idx[view1_sz:]

    # filter tokens for each view
    return ((x[idx1,:], p[idx1,:]), (x[idx2,:], p[idx2,:]))


def crop(x, p):
    # randomly choose an anchor as center of the crop
    anchor_id = torch.randint(high=len(p))
    anchor = p[anchor_id, :]

    # randomly select crop area and aspect ratio
    crop_area = torch.randint(low=crop_min, high=crop_max)
    aspect = torch.FloatTensor(1).uniform_(aspect_min, aspect_max)

    # compute min and max coordinate of the crop
    height = torch.sqrt(areas / aspect)
    width = (height * aspect)
    r0, r1 = anchor[0] - height / 2, anchor[0] + height / 2
    c0, c1 = anchor[1] - width / 2, anchor[1] + width / 2

    # filter tokens
    idx = ((p[:,0] > r0) & (p[:,0] < r1) &
           (p[:,1] > c0) & (p[:,1] < c1))
    return x[idx,:], p[idx,:]


def mask(x, p):
    # randomly select a number of tokens to keep
    mask_ratio = torch.randint(low=mask_min, high=mask_max)
    size = mask_ratio * len(p)

    # optional: set a upper bound on number of tokens
    size = torch.minimum(size, max_token_lim)

    # randomly generate index and filter tokens
    idx = torch.randperm(len(p))[:sizes]
    return x[idx,:], p[idx,:]
\end{lstlisting}
\end{algorithm}

\subsection{Relative positional embedding}\label{sec:supp.methods.position}
\begin{figure}[htb!]
    {\centering
    \includegraphics[width=0.5\textwidth]{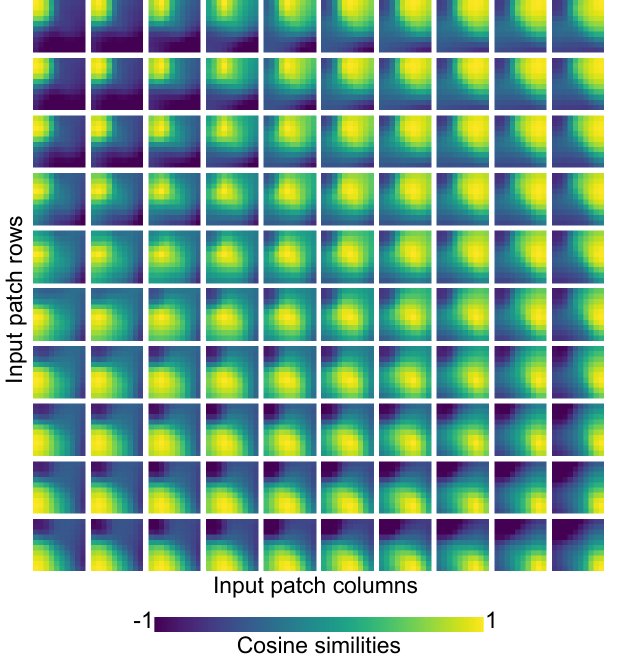}
    \caption{\textbf{Cosine similarity of learned positional embeddings.} Visualization is generated similarly as in \cite{dosovitskiy2020image}: each patch shows the cosine similarity between the position embedding of the token with the indicated row and column and the position embeddings of all other tokens.}
    \label{fig:supp.position}}
\end{figure}

To work with WSIs of different sizes, it is non-trivial to use a fixed-size learnable position embedding with absolute distance \cite{devlin2018bert}. We adopt the relative positional encoding introduced in \cite{li2021learnable}, which utilizes the learnable Fourier feature, modulated with a multi-layer perceptron. This positional encoding uses the coordinates of the patch encoder as input and the output is added directly to the transformer with the same dimension. We visualize the positional embedding similarity in Figure \ref{fig:supp.position}.

\clearpage\section{Extended Results}\label{app:ext:result}

In this section, we present extended results presented in section \ref{sec:main.results}.
\subsection{Extended ssSPT evaluation with \emph{k}NN classifier}

We evaluate ssSPT and baselines using a $k$NN classifier in Table \ref{tab:main.results.ssspt.knn}. Error bars from the table are reported in Table \ref{tab:supp.results.ssspt.knn}.

\begin{table*}[h!]
    \centering
    
    \begin{tabular}{ccccc}\toprule
    Benchmark   & Method   &    \metricheaders\\\midrule
    \multirow{4}{*}{SRH CNS}     & Mean pooling &     72.5              & 73.2         &     94.8       \\
                                 & Max pooling  &     73.5              & 75.6         &     95.0       \\
                                 & Giga-SSL     &     71.2 (0.9)        & 72.7 (0.5)   &     94.2 (1.4) \\
                                 & ssSPT (Ours) & \tb{82.3 (0.5)}   & \tb{82.3 (0.3)}  & \tb{94.7 (0.1)} \\\midrule
    \multirow{4}{*}{H\&E Glioma} & Mean pooling &     68.3       &      67.9           &     87.9       \\
                                 & Max pooling  &     68.2       &      69.2           &     87.0       \\
                                 & Giga-SSL     &     71.6 (1.3) &      71.5 (1.1)     &     89.1 (0.2) \\
                                 & ssSPT (Ours) & \tb{76.5 (0.5)} & \tb{76.1 (0.4)}    & \tb{90.9 (0.4)} \\\midrule
    \multirow{5}{*}{TCGA BRCA}   & Mean pooling &     58.1 (4.0) &      27.4 (11.3)    &     74.7 (7.9) \\ 
                                 & Max pooling  &     58.3 (3.8) &      29.1 (10.1)    &     77.6 (10.1)\\ 
                                 & HIPT         &     63.2 (6.1) &      39.9 (13.7)    &     77.5 (4.2) \\ 
                                 & Giga-SSL     &     70.4 (9.7) &      53.9 (17.5)    & \tb{85.4 (6.7)} \\ 
                                 & ssSPT (Ours) & \tb{72.7 (4.9)} & \tb{58.3 (6.8)}    &     80.4 (7.8) \\\midrule 
    \multirow{5}{*}{TCGA NSCLC}  & Mean pooling &     75.1 (3.7) &      76.5 (3.4)     &     85.3 (2.9) \\ 
                                 & Max pooling  &     79.3 (3.9) &      80.7 (3.2)     &     88.4 (1.7) \\        
                                 & HIPT         &     80.6 (2.9) &      80.1 (3.3)     &     88.9 (2.7) \\ 
                                 & Giga-SSL     &     84.3 (3.0) &      84.9 (2.6)     & \tb{92.3 (2.7)} \\ 
                                 & ssSPT (Ours) & \tb{86.1 (2.3)} & \tb{86.1 (2.6)}    & \tb{92.3 (1.7)} \\\midrule 
    \multirow{5}{*}{TCGA RCC}    & Mean pooling &     84.2 (3.3) &      86.3 (2.8)     &     96.6 (1.0) \\
                                 & Max pooling  &     79.4 (4.5) &      82.8 (3.7)     &     96.6 (1.7) \\
                                 & HIPT         &     88.4 (3.2) &      89.2 (3.0)     &     97.4 (1.6) \\
                                 & Giga-SSL     &     90.0 (2.5) &      89.3 (2.9)     &     97.5 (1.0) \\
                                 & ssSPT (Ours) & \tb{91.7 (2.1)} & \tb{90.5 (2.7)}    & \tb{98.3 (1.3)} \\\bottomrule
    \end{tabular}
    
    \caption{\textbf{Self-supervised benchmarks.} We use kNN classifier to evaluate ssSPT and baselines. Extended Table \ref{tab:main.results.ssspt.knn} with standard deviations reported in (parentheses).} 
    \label{tab:supp.results.ssspt.knn}
    \end{table*}

\subsection{Extended ssSPT and suSPT evaluation with linear classifier}

We evaluate ssSPT, suSPT, and baselines using linear evaluation in Table \ref{tab:main.results.suspt.linear}. KNN evaluation results and error bars from the table are reported in Table \ref{tab:supp.results.suspt}.

\begin{table*}[p!]
    \centering{\resizebox{\textwidth}{!}{
        
    \begin{tabular}{cccccccc}\toprule
        && \multicolumn{3}{c}{\emph{k}NN} & \multicolumn{3}{c}{Linear evaluation}\\\cmidrule(lr){3-5}\cmidrule(lr){6-8}
    Benchmark   & Method   &    \metricheaders & \metricheaders\\\midrule
    \multirow{8}{*}{SRH CNS}     & \multicolumn{7}{c}{\cellcolor{violet2}\emph{Self-supervised methods}}\\\addlinespace
                                 & Giga-SSL       &     71.2 (0.9)  &     72.7 (0.5)  &     94.2 (1.4)  &     78.6 (1.8)  &     75.6 (2.1)  &     96.1 (0.3)\\
                                 & ssSPT (Ours)   & \tb{82.3 (0.5)} & \tb{82.3 (0.3)} & \tb{94.7 (0.1)} & \tb{85.4 (0.3)} & \tb{83.2 (0.3)} & \tb{97.6 (0.3)}\\\addlinespace
                                 & \multicolumn{7}{c}{\cellcolor{orange2}\emph{Supervised methods}}\\\addlinespace
                                 & ABMIL          &     77.0 (0.9)  &     76.3 (0.1)  &     93.1 (0.6)  &     79.8 (0.4)  &     78.9 (1.0)  &     96.0 (0.4)\\
                                 & CLAM           &     -           &     -           &     -           &     83.5 (1.3)  &     82.9 (1.2)  &     96.7 (0.3)\\
                                 & TransMIL       &     84.8 (0.8)  & \tb{84.7 (0.3)} &     92.2 (0.1)  &     84.4 (1.2)  &     84.0 (0.7)  &     84.0 (0.7)\\
                                 & suSPT (Ours)   & \tb{85.0 (0.5)} &     84.2 (0.6)  & \tb{92.8 (0.7)} & \tb{85.6 (0.5)} & \tb{84.4 (0.5)} & \tb{97.2 (0.3)}\\\midrule 
    \multirow{8}{*}{H\&E Glioma} & \multicolumn{7}{c}{\cellcolor{violet2}\emph{Self-supervised methods}}\\\addlinespace
                                 & Giga-SSL       &     71.6 (1.3)  &     71.5 (1.1)  &     89.1 (0.2)  &     76.5 (0.6)  &     73.9 (0.4)  &     90.7 (0.3)\\
                                 & ssSPT (Ours)   & \tb{76.5 (0.5)} & \tb{76.1 (0.4)} & \tb{90.9 (0.4)} & \tb{80.0 (2.4)} & \tb{78.0 (2.2)} & \tb{91.9 (0.8)}\\\addlinespace
                                 & \multicolumn{7}{c}{\cellcolor{orange2}\emph{Supervised methods}}\\\addlinespace
                                 & ABMIL          &     75.6 (1.8)  &     75.7 (1.8)  & \tb{88.4 (0.4)} &     75.0 (0.6)  &     74.5 (0.6)  &     90.2 (0.3)\\
                                 & CLAM           &     -           &     -           &     -           &     78.9 (2.2)  &     77.7 (1.8)  &     88.5 (1.5)\\
                                 & TransMIL       &     79.1 (2.5)  &     78.8 (2.7)  &     86.5 (0.3)  &     80.0 (1.4)  &     79.4 (1.8)  &     91.9 (1.1)\\
                                 & suSPT (Ours)   & \tb{80.5 (1.6)} & \tb{79.9 (1.4)} &     88.3 (0.5)  & \tb{80.7 (2.4)} & \tb{80.3 (1.9)} & \tb{93.7 (0.3)}\\\midrule 
    \multirow{10}{*}{TCGA BRCA}  & \multicolumn{7}{c}{\cellcolor{violet2}\emph{Self-supervised methods}}\\\addlinespace
                                 & Giga-SSL       &     70.4 (9.7)  &     53.9 (17.5) &     85.4 (6.7)  &     81.7 (6.5)  &     62.6 (9.2)  & \tb{91.2 (5.4)}\\
                                 & ssSPT (Ours)   & \tb{72.7 (4.9)} & \tb{58.3 (6.8)} & \tb{80.4 (7.8)} & \tb{83.0 (7.3)} & \tb{67.4 (9.9)} &     89.4 (6.3)\\\addlinespace
                                 & \multicolumn{7}{c}{\cellcolor{orange2}\emph{Supervised methods}}\\\addlinespace
                                 & ABMIL          &     71.7 (9.7)  &     56.0 (17.5) &     84.9 (9.1)  &     78.6 (9.1)  &     62.8 (14.3) &     86.1 (7.5)\\
                                 & CLAM           &     -           &     -           &     -           &     76.8 (8.9)  &     59.3 (13.8) &     85.8 (6.3)\\
                                 & DSMIL          &     -           &     -           &     -           &     77.0 (6.5)  &     58.0 (7.8)  &     83.8 (7.0)\\
                                 & TransMIL       & \tb{79.0 (8.1)} & \tb{67.7 (11.2)}&     84.7 (6.8)  &     80.8 (8.1)  &     69.4 (11.3) &     90.0 (5.5)\\
                                 & HIPT           &     -           &     -           &     -           &     75.8 (8.2)  &     60.2 (11.7) &     87.4 (5.7)\\
                                 & suSPT (Ours)   &     78.2 (9.4)  &     66.1 (17.1) & \tb{86.1 (6.1)} & \tb{84.2 (5.3)} & \tb{70.4 (8.2)} & \tb{90.5 (4.8)}\\\midrule 
    \multirow{10}{*}{TCGA NSCLC} & \multicolumn{7}{c}{\cellcolor{violet2}\emph{Self-supervised methods}}\\\addlinespace
                                 & Giga-SSL       &     84.3 (3.0)  &     84.9 (2.6)  & \tb{92.3 (2.7)} &     87.5 (2.3)  &     87.9 (2.2)  &     93.7 (2.2)\\
                                 & ssSPT (Ours)   & \tb{86.1 (2.3)} & \tb{86.1 (2.6)} & \tb{92.3 (1.7)} & \tb{88.0 (3.1)} & \tb{88.0 (3.2)} & \tb{94.8 (2.1)}\\\addlinespace
                                 & \multicolumn{7}{c}{\cellcolor{orange2}\emph{Supervised methods}}\\\addlinespace
                                 & ABMIL          &     80.1 (4.0)  &     79.6 (4.5)  &     89.7 (2.1)  &     86.8 (2.4)  &     86.7 (2.3)  &     93.7 (1.2)\\
                                 & CLAM           &     -           &     -           &     -           &     84.3 (3.4)  &     83.6 (4.1)  &     92.8 (2.0)\\
                                 & DSMIL          &     -           &     -           &     -           &     84.9 (3.8)  &     84.0 (4.6)  &     92.0 (2.3)\\
                                 & TransMIL       &     88.8 (2.4)  &     88.9 (2.5)  & \tb{94.2 (2.6)} &     87.4 (2.8)  &     87.5 (2.8)  &     95.3 (1.5)\\
                                 & HIPT           &     -           &     -           &     -           &     88.2 (2.5)  &     88.2 (2.5)  &     95.2 (2.0)\\
                                 & suSPT (Ours)   & \tb{90.3 (1.8)} & \tb{90.4 (2.0)} &     94.1 (2.6)  & \tb{90.3 (2.1)} & \tb{90.4 (2.2)} & \tb{95.7 (2.4)}\\\midrule 
    \multirow{10}{*}{TCGA RCC}   & \multicolumn{7}{c}{\cellcolor{violet2}\emph{Self-supervised methods}}\\\addlinespace
                                 & Giga-SSL       &     90.0 (2.5)  &     89.3 (2.9)  &     97.5 (1.0)  &     89.9 (2.7)  &     87.8 (2.6)  &     97.7 (0.9)\\
                                 & ssSPT (Ours)   & \tb{91.7 (2.1)} & \tb{90.5 (2.7)} & \tb{98.3 (1.3)} & \tb{93.4 (1.7)} & \tb{91.1 (2.1)} & \tb{98.7 (1.1)}\\\addlinespace
                                 & \multicolumn{7}{c}{\cellcolor{orange2}\emph{Supervised methods}}\\\addlinespace
                                 & ABMIL          &     89.5 (6.0)  &     90.6 (5.2)  & \tb{97.5 (2.2)} &     89.9 (2.6)  &     89.4 (3.2)  &     97.8 (1.1)\\
                                 & CLAM           &     -           &     -           &     -           &     85.6 (8.0)  &     83.6 (7.0)  &     97.3 (1.7)\\
                                 & DSMIL          &     -           &     -           &     -           &     90.1 (3.7)  &     87.1 (4.4)  &     97.1 (1.5)\\
                                 & TransMIL       &     90.6 (2.3)  &     90.0 (2.5)  &     96.6 (1.3)  &     87.4 (2.8)  &     87.5 (2.8)  &     95.3 (1.5)\\
                                 & HIPT           &     -           &      -          &     -           &     92.0 (2.0)  & \tb{92.1 (2.6)} &     98.0 (1.2)\\
                                 & suSPT (Ours)   & \tb{92.1 (3.0)} & \tb{91.1 (3.5)} &     96.3 (2.1)  & \tb{92.3 (2.9)} &     91.0 (3.5)  & \tb{98.6 (0.7)}\\\midrule 
    \end{tabular}
    }}
    \caption{\textbf{SPT benchmarks.} Extended Table \ref{tab:main.results.suspt.linear} with KNN evaluation results and standard deviations. Standard deviations are reported in (parentheses).} 
    \label{tab:supp.results.suspt}
    \end{table*}

\clearpage

\subsection{ssSPT results with different patch encoders}

\begin{figure}[h]
    \centering
    \includegraphics[width=.5\textwidth]{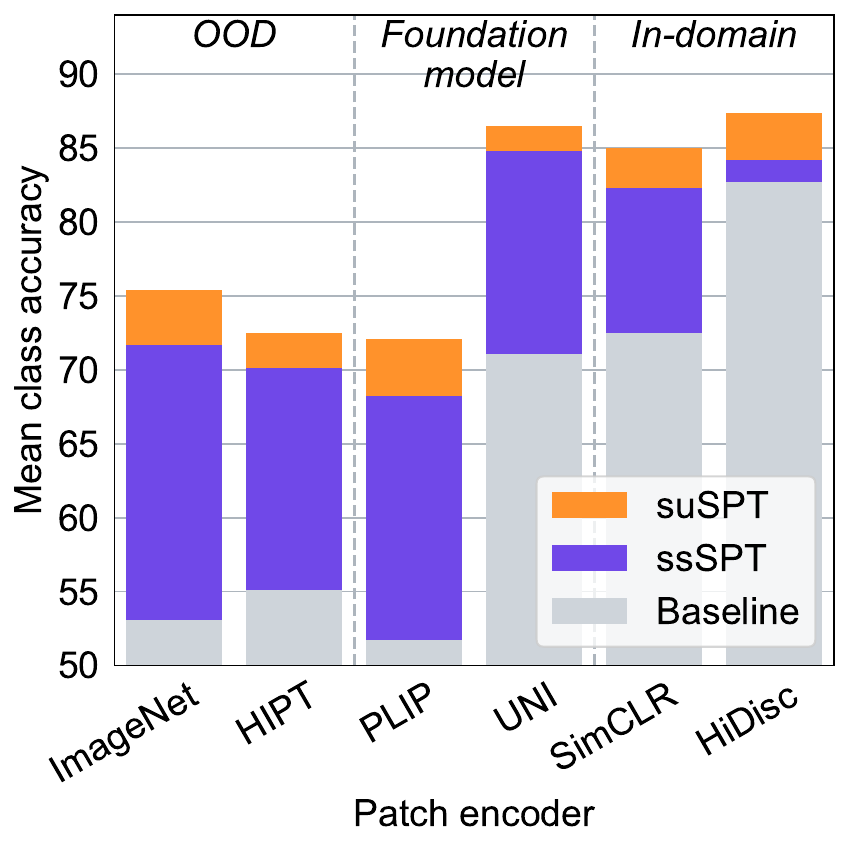}
    \caption{\tb{SPT results on SRH CNS with a wide range of patch encoders.} Publicly available foundation patch encoders are trained using H\&E data and are out-of-distribution (OOD) for SRH. In-domain patch encoders are trained using SRH data.}
    \label{fig:supp.results.patch_enc.srh}
\end{figure}

In Figure \ref{fig:main.results.patch_enc}, we compared the ssSPT's performance to the pooling baseline and suSPT performance upper bound with the H\&E glioma benchmark. Here, we present the same experiment with the SRH CNS benchmark in Figure \ref{fig:supp.results.patch_enc.srh}. In the SRH benchmark, WSI representations improved dramatically in OOD and foundation patch encoders, which are not trained on SRH data. After training with SPT, as demonstrated by a maximum 18.6\% boost in MCA for ImageNet. We also observe the expected performance boost for in-domain patch encoder SimCLR and HiDisc. Similar to the H\&E glioma benchmark, ssSPT performance approaches suSPT upperbound on all patch encoders, within 4 points in MCA for SRH CNS benchmarks.

In addition, full metrics for each patch encoder for both SRH CNS and H\&E glioma are in Table \ref{tab:supp.results.patch_encoders}.

\begin{table*}[h!]
    \centering{\resizebox{\textwidth}{!}{
    \begin{tabular}{cccccccc}\toprule
    && \multicolumn{3}{c}{SRH} & \multicolumn{3}{c}{H\&E Gliomas}\\\cmidrule(lr){3-5}\cmidrule(lr){6-8}
    Patch Encoder & Method         & \metricheaders                       & \metricheaders \\\midrule
                  & Pooling        & 53.1       & 53.0       & 88.1       & 59.8       & 74.2       & 81.7\\
    ImageNet      & ssSPT          & 71.7 (0.7) & 73.2 (0.4) & 93.7 (0.3) & 75.3 (2.5) & 74.0 (2.5) & 89.9 (0.9)\\
                  & suSPT          & 75.4 (1.4) & 76.1 (0.8) & 92.3 (0.5) & 76.4 (1.1) & 76.4 (1.2) & 88.8 (0.6)\\\midrule
                  & Pooling        & 82.7       & 82.9      &  95.9       & 74.9       & 74.2       & 90.4\\
    HiDisc        & ssSPT          & 84.2 (1.4) & 83.9 (1.2)& 95.6 (0.4)  & 77.5 (1.3) & 77.1 (1.3) & 91.2 (0.3)\\
                  & suSPT          & 87.4 (0.8) & 87.5 (0.7)& 93.3 (0.3)  & 83.6 (0.3) & 82.4 (0.2) & 88.3 (0.3)\\\midrule
                  & Pooling        & 55.1       & 56.4       & 86.6       & 54.7       & 55.0       & 77.7\\
    HIPT          & ssSPT          & 70.1 (0.4) & 71.9 (0.5) & 92.2 (0.4) & 74.9 (1.7) & 73.9 (1.7) & 89.2 (0.2)\\
                  & suSPT          & 72.5 (1.1) & 73.7 (0.9) & 90.9 (0.3) & 76.6 (1.8) & 76.1 (1.6) & 90.6 (0.7)\\\midrule
                  & Pooling        & 51.7       & 51.9       & 86.8       & 66.8       & 66.3       & 88.0\\
    PLIP          & ssSPT          & 68.2 (0.4) & 69.7 (0.4) & 92.0 (0.4) & 76.2 (2.0) & 75.5 (2.2) & 91.4 (0.4) \\
                  & suSPT          & 72.1 (1.4) & 72.6 (1.2) & 91.3 (0.4) & 82.6 (0.7) & 81.5 (0.9) & 92.5 (0.2) \\\midrule
                  & Pooling        & 71.1       & 73.4       & 93.6       & 80.3       & 78.3       & 93.5\\
    UNI           & ssSPT          & 84.8 (1.7) & 85.7 (1.5) & 96.2 (0.3) & 82.8 (2.1) & 82.3 (2.2) & 94.2 (0.5)  \\        
                  & suSPT          & 86.5 (0.9) & 86.9 (0.6) & 92.4 (0.5) & 89.1 (0.9) & 88.8 (0.7) & 92.5 (0.5)\\\bottomrule
    \end{tabular}
    }}
    \caption{\textbf{SPT results with different patch encoder.} Metrics reported for results in Figure \ref{fig:main.results.patch_enc}. Standard deviations are in (parentheses).}
    \label{tab:supp.results.patch_encoders}
\end{table*}

\subsection{suSPT benchmarks with UNI patch features}
We evaluated suSPT with UNI patch features in Table \ref{tab:main.results.suspt.uni}. Error bars are reported in Table \ref{tab:supp.results.suspt.uni}.
\begin{table*}[h!]
    \centering{\resizebox{\textwidth}{!}{
    \begin{tabular}{cccccccc}\toprule
    && \multicolumn{3}{c}{\emph{k}NN} & \multicolumn{3}{c}{Linear evaluation}\\\cmidrule(lr){3-5}\cmidrule(lr){6-8}
    Benchmark & Method       & \metricheaders                                              & \metricheaders \\\midrule
\multirow{2}{*}{SRH CNS}     & ABMIL &     84.1 (0.9)  &     84.7 (0.8)  & \tb{95.4 (0.2)} &     86.3 (0.9)  &     86.3 (0.7)  &     96.3 (0.0)  \\
                             & suSPT & \tb{86.5 (0.9)} & \tb{86.9 (0.6)} &     92.4 (0.5)  & \tb{86.4 (0.8)} & \tb{86.8 (0.5)} & \tb{98.4 (0.3)} \\\midrule
\multirow{2}{*}{H\&E Glioma} & ABMIL &     86.4 (1.5)  &     85.6 (1.4)  & \tb{94.7 (1.1)} &     86.8 (0.6)  &     85.2 (0.8)  &     95.5 (0.6)  \\
                             & suSPT & \tb{89.1 (0.9)} & \tb{88.8 (0.7)} &     92.5 (0.5)  & \tb{89.1 (1.0)} & \tb{88.7 (0.7)} & \tb{97.3 (0.3)} \\\midrule
\multirow{2}{*}{TCGA BRCA}   & ABMIL &     85.7 (6.8)  &     78.8 (9.5)  & \tb{91.7 (4.7)} &     88.6 (6.3)  &     79.1 (6.5)  &     88.8 (7.6)  \\
                             & suSPT & \tb{90.1 (6.4)} & \tb{84.2 (9.8)} &     90.3 (5.8)  & \tb{89.6 (6.3)} & \tb{82.2 (9.2)} & \tb{94.7 (3.5)} \\\midrule
\multirow{2}{*}{TCGA NSCLC}  & ABMIL & \tb{95.6 (1.2)} & \tb{95.7 (1.1)} & \tb{97.3 (2.0)} &     93.6 (1.7)  &     90.6 (2.0)  &     93.4 (2.9)  \\
                             & suSPT &     95.4 (2.4)  &     95.5 (2.3)  &     95.9 (1.8)  & \tb{95.5 (2.2)} & \tb{95.6 (2.2)} & \tb{97.6 (2.1)} \\\midrule
\multirow{2}{*}{TCGA RCC}    & ABMIL &     95.4 (1.3)  &     93.4 (2.4)  & \tb{98.3 (0.7)} &     95.2 (1.1)  &     93.3 (2.0)  &     98.5 (0.9)  \\
                             & suSPT & \tb{96.3 (2.6)} & \tb{94.9 (2.9)} &     97.2 (1.7)  & \tb{96.2 (2.8)} & \tb{94.9 (3.1)} & \tb{98.9 (0.6)} \\\midrule
    \end{tabular}
    }}
    \caption{\textbf{suSPT benchmark with UNI features.}  Extended results for Table \ref{tab:main.results.suspt.uni} with standard deviations reported in (parentheses).}
    \label{tab:supp.results.suspt.uni}
\end{table*}

\clearpage
\subsection{Ablation stuides} \label{sec:supp.results.ablation}
In this section, we present ablation studies over SSL paradigms, SPT transformation choices, and SPT transformation parameters.

\subsubsection{SSL paradigms}
We examine the effect of different SSL paradigms on ssSPT in table \ref{tab:supp.results.ssspt.ablation.algo}.
VICReg performs the best on the SRH CNS benchmark, BYOL performs the best on the TCGA RCC benchmark, while SimCLR performs the best on H\&E glioma, TCGA BRCA, and TCGA NSCLC.
These results demonstrate that the flexibility of SPT allows different algorithms to accommodate different training dynamics of different diagnostic tasks and image modalities.

\begin{table*}[htb!]
    \centering{\resizebox{\textwidth}{!}{
    \begin{tabular}{cccccccc}\toprule
    && \multicolumn{3}{c}{\emph{k}NN} & \multicolumn{3}{c}{Linear evaluation}\\\cmidrule(lr){3-5}\cmidrule(lr){6-8}
    Benchmark   &  Algorithm   &    \metricheaders                                  &    \metricheaders\\\midrule
                & ssSPT-SimCLR &     77.3 (0.6)  &     77.6 (0.3)  &     94.0 (0.4)  &     82.7 (0.7)  &     81.8 (0.5)  &     97.5 (0.3) \\
    SRH CNS     & ssSPT-VICReg & \tb{82.3 (0.5)} & \tb{82.3 (0.3)} & \tb{94.7 (0.1)} & \tb{85.4 (0.3)} & \tb{83.2 (0.3)} & \tb{97.6 (0.3)}\\
                & ssSPT-BYOL   &     79.4 (0.4)  &     78.7 (1.2)  &     94.3 (0.4)  &     83.4 (0.5)  &     81.0 (1.3)  &     96.9 (0.0) \\\midrule
                & ssSPT-SimCLR & \tb{76.5 (0.5)} & \tb{76.1 (0.4)} & \tb{90.9 (0.4)} & \tb{80.0 (2.4)} & \tb{78.0 (2.2)} &     91.9 (0.8) \\
    H\&E Glioma & ssSPT-VICReg &     72.8 (0.7)  &     72.4 (0.9)  &     87.9 (0.7)  &     74.5 (0.8)  &     73.1 (0.8)  &     90.9 (0.5) \\
                & ssSPT-BYOL   &     74.6 (0.9)  &     73.8 (0.9)  &     89.0 (0.5)  &     77.8 (0.6)  &     76.1 (0.8)  & \tb{92.2 (0.0)}\\\midrule
                & ssSPT-SimCLR & \tb{72.7 (4.9)} & \tb{58.3 (6.8)} &     80.4 (7.8)  &     83.0 (7.3)  & \tb{67.4 (9.9)} &     89.4 (6.3) \\
    TCGA BRCA   & ssSPT-VICReg &     69.2 (7.9)  &     51.0 (15.2) &     77.4 (9.3)  &     78.6 (5.4)  &     59.3 (7.5)  &     85.3 (7.7) \\
                & ssSPT-BYOL   &     70.0 (8.3)  &     52.6 (15.8) & \tb{81.1 (9.0)} & \tb{83.6 (5.4)} &     66.7 (6.7)  & \tb{90.4 (6.6)}\\\midrule
                & ssSPT-SimCLR & \tb{86.1 (2.3)} & \tb{86.1 (2.6)} & \tb{92.3 (1.7)} & \tb{88.0 (3.1)} & \tb{88.0 (3.2)} & \tb{94.8 (2.1)}\\
    TCGA NSCLC  & ssSPT-VICReg &     80.2 (3.4)  &     80.7 (3.0)  &     86.0 (3.4)  &     87.6 (2.2)  &     87.9 (2.0)  &     93.8 (1.6) \\
                & ssSPT-BYOL   &     81.4 (2.4)  &     81.8 (2.7)  &     87.2 (2.0)  &     86.0 (2.8)  &     86.2 (3.1)  &     92.7 (2.2) \\\midrule
                & ssSPT-SimCLR &     90.7 (2.9)  &     90.4 (2.7)  & \tb{98.3 (1.0)} &     91.7 (2.1)  &     89.4 (3.2)  &     98.2 (0.7) \\
    TCGA RCC    & ssSPT-VICReg &     87.2 (3.9)  &     87.2 (3.1)  &     96.0 (1.6)  &     88.6 (3.3)  &     86.6 (3.2)  &     96.9 (0.9) \\
                & ssSPT-BYOL   & \tb{91.7 (2.1)} & \tb{90.5 (2.7)} & \tb{98.3 (1.3)} & \tb{93.4 (1.7)} & \tb{91.1 (2.1)} & \tb{98.7 (1.1)}\\\bottomrule

    \end{tabular}}}
    
    \caption{\textbf{SPT benchmark with different SSL paradigms.} We compare different SPT training algorithms.  Standard deviations are reported in (parentheses).} 
    \label{tab:supp.results.ssspt.ablation.algo}
    \end{table*}

\subsubsection{SPT transformation choices}
Table \ref{tab:supp.results.ablate.aug} reports the performance of individual and combinations of slide-level transformations. We leave out splitting alone since it does not reduce the total number of patch tokens for both views, making it prohibitive to train. For the masking and cropping alone, we show that cropping performs better than masking for both SRH CNS and H\&E Glioma. Combining cropping and masking shows similar performance with cropping alone, but is more efficient to train due to the reduced number of tokens per view. Combining each transformation with splitting improves or maintains performance, showing the benefit of further decreasing mutual information (MI) between views.  Overall, cropping is the most important transformation, because capturing regional heterogeneity is the most effective and challenging pre-text task.
\begin{table*}[htb!]
    \centering{
    \begin{tabular}{ccccccc}\toprule
                     &&  & \multicolumn{2}{c}{SRH CNS} & \multicolumn{2}{c}{H\&E Glioma}\\
                       \cmidrule(lr){4-5}\cmidrule(lr){6-7}
    Splitting & Cropping & Masking & MCA        & F1         & MCA        & F1 \\\midrule
    &&X     & 75.4 (0.9) & 76.0 (0.7) & 69.9 (0.5) & 69.7 (0.5)\\
    &X&     & 81.3 (1.5) & 81.5 (1.5) & 74.3 (0.2) & 74.4 (0.2)\\
    X&&X    & 76.1 (0.3) & 76.3 (0.8) & 69.2 (1.5) & 69.3 (1.5)\\
    X&X&    & 82.1 (1.4) & 82.2 (1.2) & 73.6 (0.6) & 73.8 (0.4)\\
    &X&X    & 81.2 (0.5) & 81.8 (0.5) & 75.5 (2.1) & 75.4 (2.2)\\
    X&X&X   & \textbf{82.3 (0.5)} & \textbf{82.3 (0.3)} & \textbf{76.5 (0.5)} & \textbf{76.1 (0.4)}\\\bottomrule
    \end{tabular}
    }
    \caption{\textbf{Ablation over transformations.} X stands for transformations used. Standard deviations reported in (parentheses).}
    \label{tab:supp.results.ablate.aug}
\end{table*}

\subsubsection{SPT transformation parameters}
We examine the relationship between mutual information (MI) and model performance by combining and adjusting SPT transformation parameters in the H\&E glioma benchmark.

\begin{figure}[h]
     \centering

     \includegraphics[width=\textwidth]{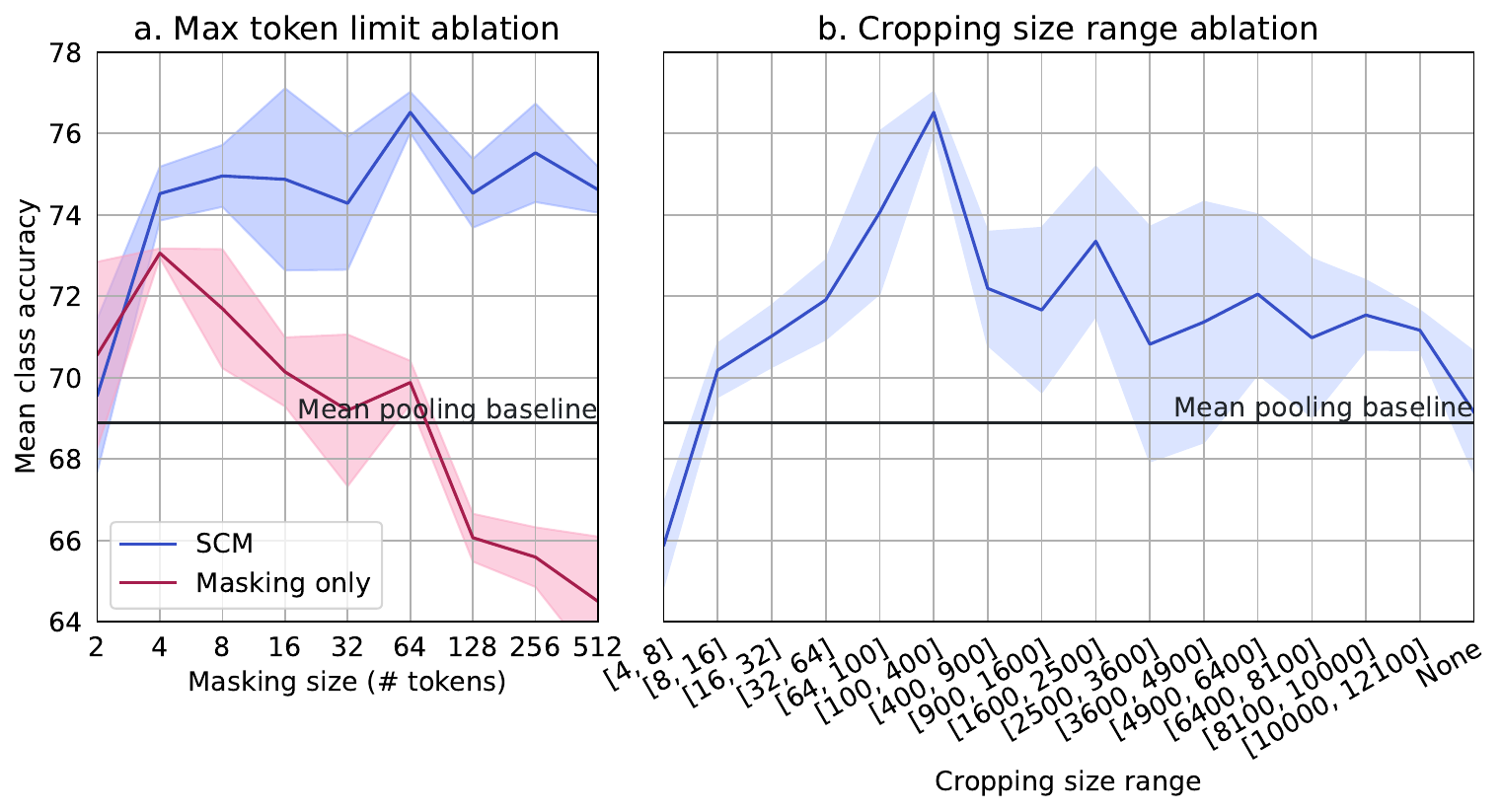}
     \caption{\textbf{SPT transformation parameter ablation for H\&E glioma molecular classification.}
     \textbf{a) Max token limit ablation.} SCM, splitting-cropping-masking transformation.
     \textbf{b) Cropping size range ablation.} Cropping size range [min, max cropping size] is on the \emph{x}-axis. None represent no cropping applied.
     The shaded area represents the standard deviation across three different random seeds. 
     }\label{fig:supp.results.ablation.aug_params}
\end{figure}

\paragraph{Token limit.}  We studied the effect of the number of tokens, with splitting + cropping + masking and only masking. A higher token limit results in a higher likelihood of overlapping area coverage between views. The results are in Figure \ref{fig:supp.results.ablation.aug_params}a. As expected, MI levels can be adjusted to optimize model performance. Previous studies have concluded that fewer patch tokens resulted in better model performance \cite{lazard2023giga}, but only in the case of using masking transformation alone.

\paragraph{Cropping size.} We varied cropping size range in Figure \ref{fig:supp.results.ablation.aug_params}b. Cropping also allows us to adjust the amount of MI between views, where a larger crop size leads to a higher probability of overlapping regions between views. Our ablation study shows that for SimCLR training on our H\&E glioma benchmark, $[100, 400]$ is the optimal cropping size range, achieving a MCA of 76.5. These results correspond to the hypothesis in \cite{tian2020makes}, where an optimal amount of MI between views achieves the best performance. 

\clearpage
\subsubsection{SPT model size}

We performed additional ablation studies with different SPT model sizes, and the results are reported in Table \ref{tab:supp.results.model_size}.
The six- and two-layer transformers have the best performance for H\&E glioma and SRH CNS, respectively. Smaller models performed better for SRH CNS, likely because of a smaller slide size and a relatively more uniform image. All models ranging from two to eight-layer transformers outperformed previous best self-supervised models.

\begin{table*}[h!]
    \centering{\resizebox{\textwidth}{!}{
    \begin{tabular}{cccccccc}\toprule
    && \multicolumn{3}{c}{SRH} & \multicolumn{3}{c}{H\&E Glioma}\\\cmidrule(lr){3-5}\cmidrule(lr){6-8}
    \# Layers & \# Parameters & \metricheaders & \metricheaders \\\midrule
    2 & 6.7 M  & \tb{83.5 (0.1)}  & \tb{83.8 (0.3)}  & \tb{95.4 (0.4)} &     74.9 (0.9)  &     74.8 (1.0)  &     91.6 (0.5)\\
    4 & 13 M   &     82.2 (0.9)   &     83.2 (1.6)   & \tb{95.4 (0.2)} &     75.5 (0.8)  &     75.3 (0.5)  &     91.1 (0.4)\\
    6 & 19.3 M &     82.3 (0.5)   &     82.3 (0.3)   &     94.7 (0.1)  & \tb{76.5 (0.5)} & \tb{76.1 (0.4)} &     90.9 (0.4)\\
    8 & 25.6 M &     81.4 (0.2)   &     82.5 (0.3)   &     95.2 (0.2)  &     75.2 (1.2)  &     75.2 (1.1)  & \tb{91.7 (0.9)}\\\bottomrule
    \end{tabular}
    }}
    \caption{\textbf{SPT ablation studies on model size.} Standard deviations are in (parentheses).}
    \label{tab:supp.results.model_size}
\end{table*}

\subsection{Extended attention heatmaps}\label{sec:supp.results.attention}
\begin{figure*}[p!]
    \centering
    \includegraphics[width=\textwidth]{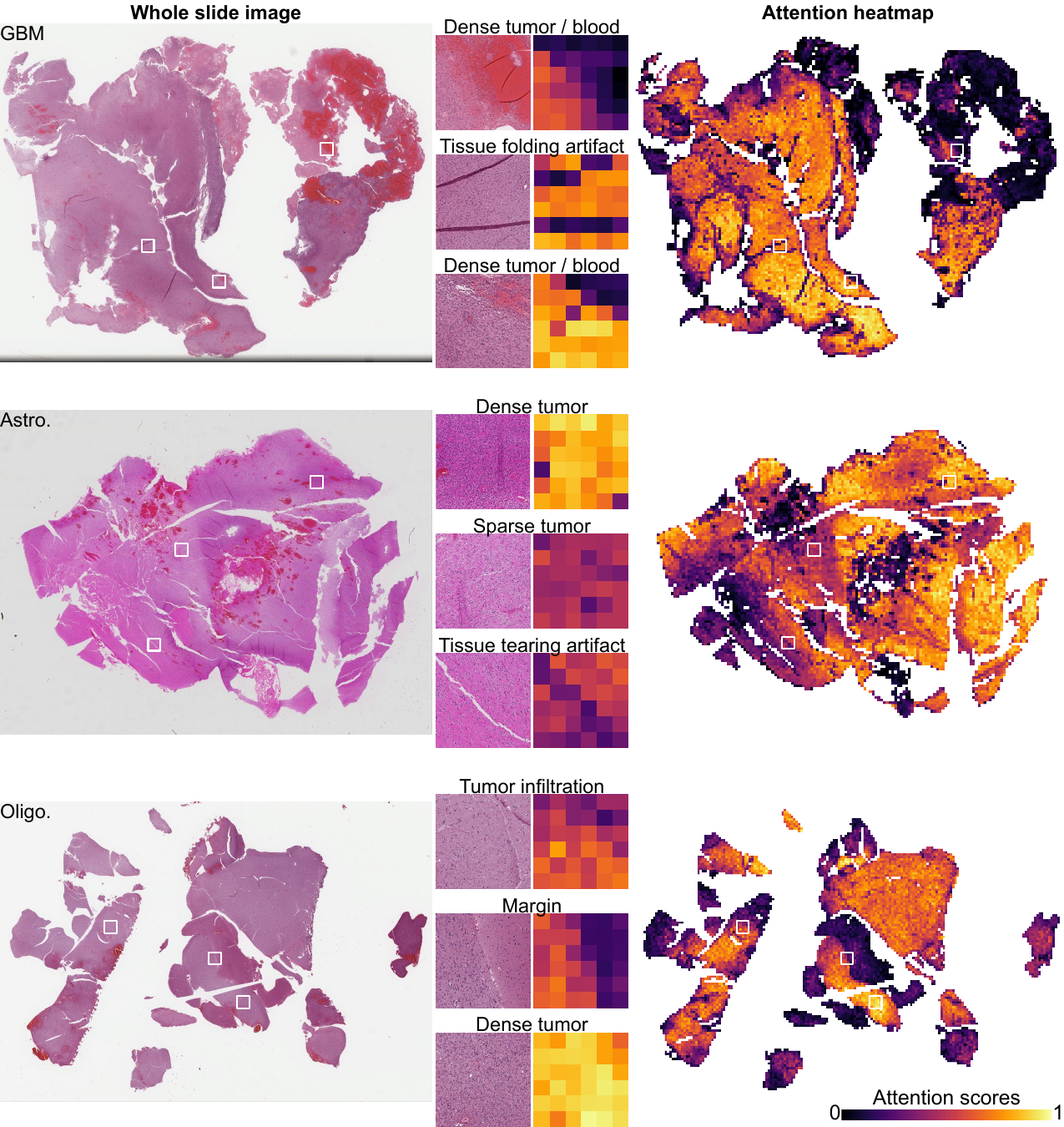}
   \caption{\textbf{Extended attention heatmap for H\&E whole slides.} The attention map shows the ssSPT-trained whole slide transformer can differentiate between different morphologies like dense tumors, blood, and different artifacts on the gigapixel WSIs. Visualization shows high attention to varying degrees of tumor infiltration, and low attention to low cellularity, blood, necrotic regions, and tissue processing artifacts. Oligo, oligodendroglioma, Astro, astrocytoma, GBM, glioblastoma.}
    \label{fig:supp.results.heatmap}
\end{figure*}

\begin{figure*}[p!]
    \centering
    \includegraphics[width=\textwidth]{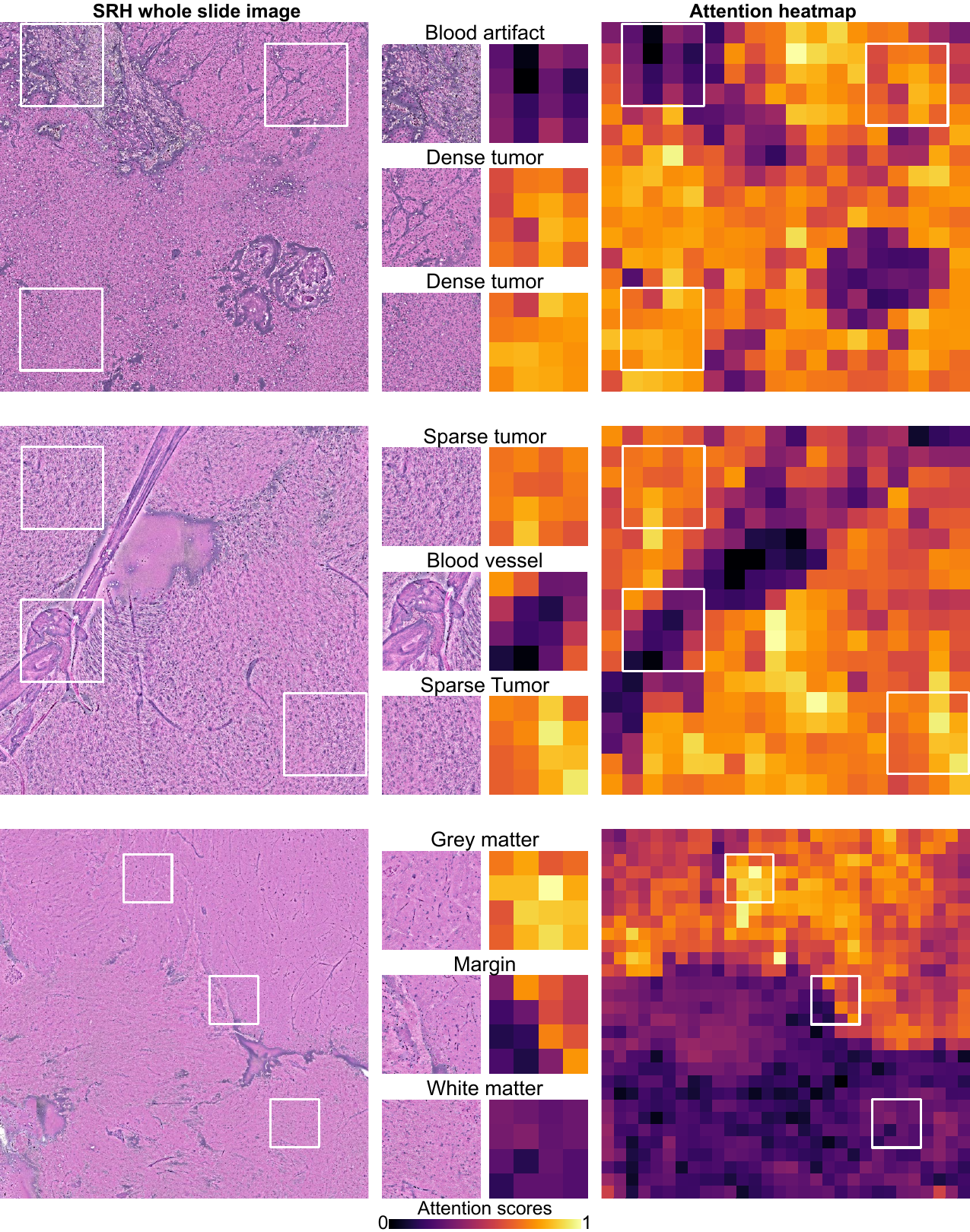}
    \caption{\textbf{Extended attention heatmap for SRH  whole slides.} On this SRH WSI, relatively small compared to H\&E, the model shows a strong capacity for unsupervised segmentation of histological features. Visualization shows high attention values to tumor regions, and low attention values to non-diagnostic regions such as blood vessels, laser noise, and empty space.}
    \label{fig:supp.results.srh_heatmap}
\end{figure*}

We present extended attention heatmap visualization for SRH and H\&E in Figures \ref{fig:supp.results.heatmap} and     \ref{fig:supp.results.srh_heatmap}, respectively. Recall these attentions are the self-attention of the CLS token from the last layer of the whole slide encoder. On both SRH and H\&E heatmaps, we can see that the whole slide encoder attends to clinically significant regions such as dense tumors, and regions with tumor infiltration, and avoids non-diagnostic regions such as blood, tissue processing artifacts, and blood vessels.

\end{document}